%% file: main.tex
\documentclass{article} 
\PassOptionsToPackage{table}{xcolor}
\usepackage[main,preprint]{mystyle}
\usepackage[utf8]{inputenc}
\usepackage[T1]{fontenc}

\input{math_commands.tex}

\usepackage{hyperref}
\usepackage{url}
\usepackage{graphicx}
\usepackage{bxcoloremoji}
\usepackage{float}
\usepackage{booktabs}
\usepackage{multirow}
\usepackage[table]{xcolor}
\usepackage{enumitem}
\usepackage{caption}
\usepackage[most]{tcolorbox}
\usepackage{wrapfig}

\usepackage{amsthm}

\AddToHook{env/table/begin}{\vspace{-6pt}}
\AddToHook{env/table/end}{\vspace{-6pt}}

\title{AdaVDR: Adaptive Tool Use and Reflection for Video Deep Research}

\author{
\textbf{Xintong Zhang}\textsuperscript{1,2,*},
\textbf{Xiaomeng Fan}\textsuperscript{2,*},
\textbf{Shilin Yan}\textsuperscript{1},
\textbf{Ekko He}\textsuperscript{1},
\textbf{Zicheng Liu}\textsuperscript{1},\\
\textbf{Zijian Zou}\textsuperscript{1},
\textbf{Guannan Zhang}\textsuperscript{1},
\textbf{Yuwei Wu}\textsuperscript{2},
\textbf{Zhi Gao}\textsuperscript{2,\textdagger},
\textbf{Hongwei Xue}\textsuperscript{1,\textdagger}\\[4pt]
\normalfont
\textsuperscript{1}Accio Team, Alibaba Group\\
\textsuperscript{2}Beijing Key Laboratory of Intelligent Information Technology,\\
School of Computer Science \& Technology, Beijing Institute of Technology\\
}

\headerlogos{
	\includegraphics[height=0.5cm]{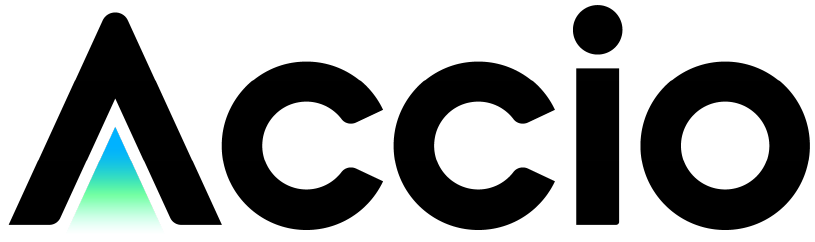}  
}

\newcommand{\github}{\raisebox{-1.5pt}{\includegraphics[height=1.05em]{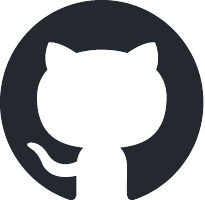}}}
\begin{document}
	
\maketitle

\insert\footins{\noindent\footnotesize $^*$Equal contribution. $^\dagger$Corresponding authors. This work was done during Xintong Zhang's internship at the Accio Team, Alibaba Group.}

\vspace{-0.5cm}

\begin{abstract}

    Video deep research aims to answer questions by jointly understanding video content and retrieving external knowledge from the open Web.
    However, these tasks often require diverse and multi-step tool-use trajectories, which may lead to unnecessary video tool calls, incurring additional costs and failure risk.
    In this paper, we propose AdaVDR, an \emph{adaptive video deep research agent} with adaptive tool use and reflection, which selects tools based on task and model capabilities while backtracking when needed. 
    To this end, we develop a data construction pipeline that builds adaptive tool-use trajectories according to task types and model capabilities. Specifically, we construct QA pairs that require both video and external information, and organize the information acquisition processes into task-specific tool-use trajectories. 
    We then apply model-conditioned tool necessity filtering to remove tools that can be bypassed using video understanding and internal knowledge of the model.
    Based on this pipeline, we construct a training dataset with adaptive trajectories and perform SFT on them to initialize these capabilities of AdaVDR.
    To further improve adaptive capabilities, we propose NASA-GRPO, a necessity-aware segment-guided credit assignment method, which evaluates segment-guided evidence contribution and tool necessity to construct normalized tool-level rewards.
    We also construct VDR-EE, a benchmark built using the proposed pipeline. 
    Experiments show that AdaVDR outperforms the base model by 15.20 and 29.00 percentage points on VDR-EE and the existing VideoDR, respectively, demonstrating its superiority.

\par\vspace{0.3cm}
\noindent\coloremojicode{1F310}  \textbf{Project Page:} \href{https://Accio-Lab.github.io/AdaVDR}{https://Accio-Lab.github.io/AdaVDR} \vspace{0.1cm}

    \github{}  \textbf{Github Repo:} \href{https://github.com/Accio-Lab/AdaVDR}{https://github.com/Accio-Lab/AdaVDR}

\end{abstract}

\section{Introduction}
\vspace{-2pt}

\begin{figure}[!t]
    \centering
    \IfFileExists{figs/teaser.pdf}{
        \includegraphics[width=\linewidth]{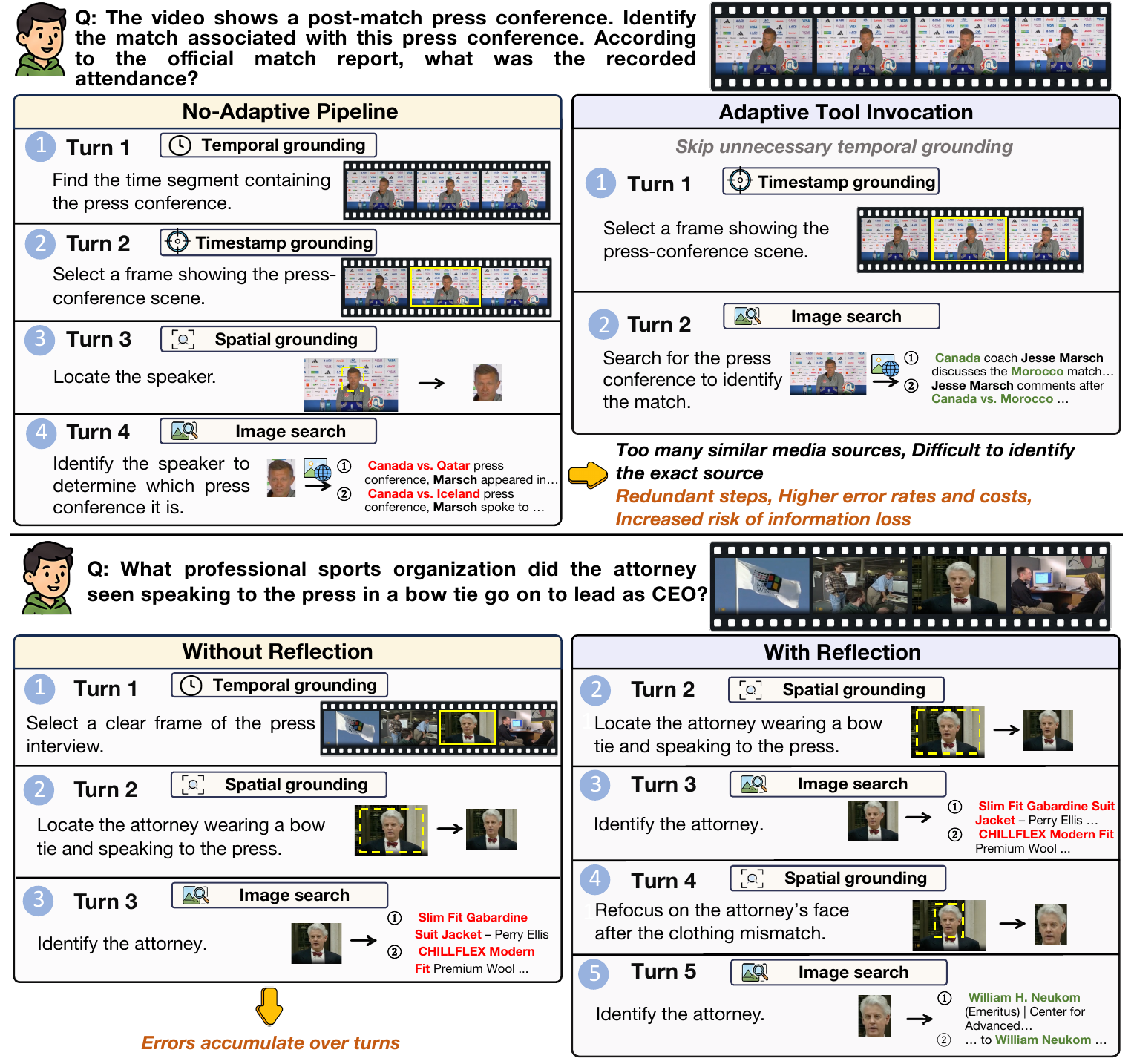}
    }{
        \fbox{\rule{0pt}{2in}\rule{0.98\linewidth}{0pt}}
    }
    \vspace{-6pt}
    \caption{\textbf{Adaptive tool use and reflection in AdaVDR.} AdaVDR adaptively selects tools and reflects on unreliable evidence, reducing redundant calls and error propagation.}
    \label{fig:teaser}
    \vspace{-9pt}
\end{figure}

Video deep research is an important task that aims to answer complex questions by jointly reasoning over video content and external knowledge. 
Solving such tasks requires not only understanding the video, but also retrieving, verifying, and integrating relevant information from the Web~\citep{liu2026watching,gao2026videosearcher,liang2025video}.
A video deep research agent invokes video understanding tools, such as grounding, and retrieval tools, such as image search, to progressively acquire evidence and complete the task~\citep{zhang2026deep,li2026videothinker,wu2025mmsearch}.

A key challenge in video deep research is that unnecessary or inappropriate tool calls lead to inefficient costs and incorrect answers during multi-step tool use. 
First, diverse questions and videos require different tool-use trajectories, and inappropriate tool use may fail to acquire critical evidence~\citep{yuan2025videoexplorer,li2026videothinker}. Second, video grounding and external retrieval tools are inherently error-prone~\citep{zhi2025videoagent2,zuo2026videolucy,liu2026watching}, so unnecessary calls can introduce erroneous information and disrupt subsequent process~\citep{li2026hypereyes,chu2026redsearcher}. Our theoretical analysis further shows additional tool calls increase prediction risk when tools are unreliable. These observations motivate us to adaptively select necessary tools according to task requirements, model capabilities, and internal knowledge.

In this paper, we propose AdaVDR, as illustrated in Fig.~\ref{fig:teaser}, an \emph{adaptive video deep research agent} with adaptive tool invocation and reflection.
For tool invocation, the agent selects necessary tools according to the question type, its video understanding capability, and internal knowledge.
For example, AdaVDR skips temporal grounding when the relevant frame can be directly located, and image search when the entity can be recognized.
For reflection, AdaVDR backtracks when intermediate evidence is unreliable, re-performing according to the failure. AdaVDR reduces redundant tool use and prevent unreliable evidence from affecting subsequent processes.



To equip AdaVDR with the adaptive tool-use capability, we propose a video deep research data construction pipeline consisting of two stages: \emph{QA generation} and \emph{trajectory generation}. 
For QA generation, we identify events and entities in videos that require external retrieval, and combine grounding and retrieval to acquire relevant information and construct QA pairs that rely on both video content and external knowledge. For trajectory generation, we organize and refine the grounding and retrieval processes into tool-use trajectories according to the information needs of different questions. We then perform tool necessity filtering based on video understanding capability and internal knowledge of the target model, removing tools that the model can bypass to obtain adaptive training trajectories tailored to its capabilities.
Based on the pipeline, we construct a training dataset and a benchmark, VDR-EE, which includes fine-grained annotations of question types.

We perform supervised fine-tuning (SFT) on the constructed adaptive trajectories to equip AdaVDR with initial adaptive capabilities.
To  further improve adaptive capabilities of AdaVDR, we propose NASA-GRPO, a necessity-aware segment-guided credit assignment method, which assigns tool-level rewards to tool calls according to their importance to strengthen the adaptive ability. NASA-GRPO evaluates segment-guided evidence contribution and tool necessity, and applies tool-type reward normalization to construct tool-level rewards.
We evaluate AdaVDR on both the constructed VDR-EE benchmark and the existing VideoDR benchmark.
Our method outperforms the baselines by 15.20 and 29.00 percentage points on VDR-EE and VideoDR, respectively, demonstrating its superiority. Results also demonstrate that our agent effectively reduces redundant tool calls, leading to improved performance and lower inference latency. The contributions can be summarized as four-fold.

\begin{itemize}
    \item  We propose AdaVDR, an adaptive video deep research agent that selects tools according to task requirements and model capabilities, and selectively reflects on unreliable intermediate evidence.
    \item We develop a video deep research data construction pipeline that generates task-specific tool-use trajectories and filters unnecessary tools according to target model capabilities.
    \item  We provide training data and a benchmark with fine-grained question-type annotations to support future research.
    \item We propose NASA-GRPO, a necessity-aware segment-guided credit assignment method that computes tool-level rewards for tool importance and improves necessary tool identification.
\end{itemize}

\vspace{-2pt}
\section{Related Work}

\subsection{Video Understanding Agents}
Video understanding agents improve long-video reasoning by iteratively locating question-relevant evidence within videos. Representative systems such as VideoMind \citep{liu2025videomind}, LVAgent \citep{chen2025lvagent}, VCA \citep{yang2025vca}, and Deep Video Discovery \citep{zhang2026deep} identify relevant segments, frames, actions, or entities through agentic video exploration. Recent approaches further improve this process through uncertainty-aware plan adjustment \citep{zhi2025videoagent2}, hierarchical memory backtracking \citep{zuo2026videolucy}, parallel segment-level perception and global aggregation \citep{pang2025mr}, or dynamic temporal grounding guided by intermediate sub-questions \citep{yuan2025videoexplorer}. Their evidence acquisition primarily remains within the video, whereas Video Deep Research additionally requires external information associated with grounded entities or events. Our work studies how video grounding and external retrieval can be adaptively coordinated according to task requirements and model capabilities, while allowing the agent to revise unreliable intermediate results during multi-turn reasoning.

\subsection{Multimodal Deep Research}
Deep-search agents iteratively reason and invoke external tools to acquire information \citep{yao2022react,jin2025search,li2025websailor}. Multimodal extensions incorporate image search, visual grounding, cropping, and other visual operations, and learn tool-use policies through supervised fine-tuning or reinforcement learning \citep{wu2025mmsearch,hong2026deepeyesv2,huang2601vision,chen2026opensearch,zhang2026visual}. Related work also explores visual tool coordination \citep{gao2024clova,zhang2025adaptive,he2025rsagent}, adaptive tool use and mode selection \citep{yan2026act,zhang2026adaptmmbench}, and longer multimodal search trajectories with richer visual interactions \citep{li2026hypereyes,chu2026redsearcher}. These methods primarily operate on static input images or visual evidence retrieved from the Web. Video Deep Research introduces an additional temporal dimension: relevant evidence may concern an entity appearing at a particular moment or an event unfolding over an extended segment, requiring video grounding to be coordinated with external retrieval.

Recent studies formulate and evaluate video deep research \citep{liu2026watching} and develop agents that combine video perception with search. VideoSearcher integrates video grounding with multimodal search, VideoRover coordinates video cropping, search, and Web browsing, and Video-DeepResearch adopts a staged perception--exploration pipeline \citep{gao2026videosearcher,liu2026thinking,fang2026video}. Beyond combining video and search tools, an effective agent must determine which grounding and retrieval operations are needed for each question and whether the model can obtain the required information without invoking them. AdaVDR addresses these requirements through adaptive tool use and reflection. Its data construction adapts trajectories to task requirements and model capabilities, while NASA-GRPO combines segment-guided evidence credit with model-conditioned tool necessity to discourage redundant interactions and reduce error propagation.

\vspace{-4pt}
\section{Adaptive Video Deep Research}
\vspace{-4pt}

We first formulate video deep research as a multi-turn reasoning and evidence-acquisition task over videos and external tools, and characterize its unique challenges. We present a theoretical analysis of the risks induced by unnecessary tool use. We then introduce an adaptive agent mechanism that dynamically determines both which tools to invoke and when reflection is necessary.

\vspace{-2pt}
\subsection{Problem Definition}
Given a raw video $V$ and a user question $q$, the task of \emph{video deep research} aims to generate an accurate textual answer $a$ by iteratively reasoning and leveraging a suite of external tools $\mathcal{T} = \{T_1, T_2, \dots, T_K\}$.
The toolset encompasses three primary categories: \emph{video grounding} (including temporal grounding, timestamp grounding, and spatial grounding), \emph{external retrieval} (including image search, web search, and page visits), and \emph{termination} for final answer generation.
Formally, at each interaction turn $t$, the agent observes the interaction history $h_t=(q,v,a_1,o_1,\dots,a_{t-1},o_{t-1})$ and generates an assistant output $a_t=\pi(h_t)$, where $\pi$ is typically a VLM acting as the agent brain. At each non-terminal turn, $a_t$ jointly contains the agent's reasoning and a selected tool action from $\mathcal{T}$ with its corresponding arguments, while $o_t$ denotes the observation returned by executing that action. Once sufficient evidence has been acquired, the final assistant output instead contains the agent's reasoning and textual answer, without an additional tool observation.

Compared with \emph{image deep research}, video deep research involves more diverse question types, video content, and tool-use requirements, leading to substantially different information needs and reasoning trajectories across tasks. Two representative question types are:

\begin{itemize}[leftmargin=*]
    \item \noindent\textbf{Temporal Event.} Event-centric questions require the agent to identify an action or event and localize the corresponding temporal interval. They often rely on temporal or timestamp grounding to determine when the relevant evidence appears, while spatial grounding is typically unnecessary. Some event-related questions can further proceed to text-based web search once the video content has been sufficiently understood.

    \item \noindent\textbf{Entity.} Entity-centric questions require the agent to identify or ground an object, person, product, or other entity at a particular timestamp or across multiple frames. Such questions often require locating a relevant frame, grounding the target region, and further using OCR, image search, or web search to resolve the entity's fine-grained identity and related information.

\end{itemize}

These differences make the choice of tool-use strategy critical. \textit{An inappropriate reasoning trajectory may fail to acquire the information required by the task or introduce irrelevant operations, ultimately leading to incorrect reasoning results.}

The outputs of video grounding and external retrieval tools are inherently uncertain. Temporal or timestamp grounding may locate an inaccurate segment or frame, spatial grounding may focus on an uninformative region, and image or web search may return irrelevant results. As a result, when the required information can already be obtained from the model's video understanding capability or internal knowledge, executing additional tool calls unnecessarily exposes the reasoning process to these uncertainties.\textit{ Such redundant interactions not only increase inference latency, but may also introduce noisy or incorrect evidence and degrade the final reasoning result.}

To address these challenges, we propose an \textbf{adaptive reasoning} paradigm with \emph{adaptive tool invocation} and \emph{adaptive reflection}. For tool invocation, the agent first selects the most appropriate grounding and retrieval tools according to the question type and video content, allowing different tasks to follow different reasoning trajectories. It further adapts tool use to its own capability boundary and internal knowledge, skipping grounding or retrieval steps when the required information can already be obtained directly. Reflection is also performed adaptively: reliable intermediate results are directly used for subsequent reasoning, whereas unreliable or insufficient results trigger the agent to identify the problematic grounding or retrieval step and backtrack for correction. In this way, the agent avoids both inappropriate and unnecessary tool interactions while reducing the propagation of unreliable information.

\subsection{Theoretical Analysis}
\label{section:theoretical_analysis}

Let $X=\{V,q\}$ denote the input and $Y$ the target output.
For a model $M$, let $H_M=h_M(X)$ denote the information extracted by the model
from $X$. Let $Z$ denote the observation from an external tool. Let $C\in\{0,1\}$ indicate whether the observation is erroneous, and let
$\epsilon=\Pr(C=1)$ denote the tool error probability. 




\begin{theorem}
\label{theorem:theorem1}
We define the \emph{base risk} $R_{\mathrm{base}}$ and the \emph{tool-augmented risk} $R_{\mathrm{tool}}$ as
\begin{equation}
\small
\begin{aligned}
    R_{\mathrm{base}}
    = \mathbb{E}\left[-\log P(Y\mid H_M;M)\right],
    R_{\mathrm{tool}}
    = \mathbb{E}\left[-\log P(Y\mid H_M,Z;M)\right].
\end{aligned}
\end{equation}
The difference between the base risk $R_{\mathrm{base}}$ and the tool-augmented risk $R_{\mathrm{tool}}$ satisfies
\begin{equation}
\small
R_{\mathrm{tool}} - R_{\mathrm{base}}
\ge
2\epsilon\delta^2
-
\Gamma_M
-
\xi_M.
\label{eq:tool_harm_bound}
\end{equation}
 $\Gamma_M$  measures the necessary evidence in $X$ that is not captured by $H_M$.
 $\delta$  measures how strongly erroneous tool outputs distort the model prediction.
 $\xi_M$  measures the
mismatch between the ideal prediction and the actual prediction of model $M$
given $H_M$.
\end{theorem}

Specific definition and proof are presented in appendix. Theorem~\ref{theorem:theorem1} shows that using external tools introduce additional prediction risk, when  $2\epsilon\delta^2 > \Gamma_M + \xi_M$. 
Video deep search is characterized by a high tool error probability $\epsilon$, resulting in a large $2\epsilon\delta^2$.
When the model captures sufficient information (small $\Gamma_M$), such that $2\epsilon\delta^2 > \Gamma_M+\xi_M$, 
it tends to
$R_{\mathrm{tool}} > R_{\mathrm{base}}$. In contrast, when information remains uncaptured (large $\Gamma_M$), 
it tends to $R_{\mathrm{tool}} < R_{\mathrm{base}}$.
This motivates our adaptive tool use, which selects tools according to the requirements of each task and the capability of the target model, thereby minimizing the prediction risk.

\vspace{-2pt}
\subsection{Adaptive Agent}
\vspace{-2pt}

 We develop AdaVDR, an adaptive agent with adaptive tool invocation and reflection.
 For tool invocation, AdaVDR selects the most appropriate tool-use trajectory according to the question type.
 AdaVDR further adapts tool use to its own capability and internal knowledge, skipping grounding or retrieval steps when the information can be obtained directly.  For reflection, only unreliable results trigger AdaVDR to identify the problematic tool calls and backtrack for correction.
  Thus, AdaVDR avoids inappropriate or unnecessary tool calls and limits the spread of unreliable information.

To equip AdaVDR with such adaptive capabilities, we construct adaptive tool-use trajectories according to task types and target model capabilities (Section~\ref{section:data_collection}) and perform SFT on them to initialize these capabilities.
We introduce a novel reinforcement learning (RL) method that builds tool-level rewards based on tool contribution and necessity, further strengthening adaptive tool use (Section~\ref{section:training}).

\begin{figure}
    \centering
    \IfFileExists{figs/data_collection.pdf}{
        \includegraphics[width=\linewidth]{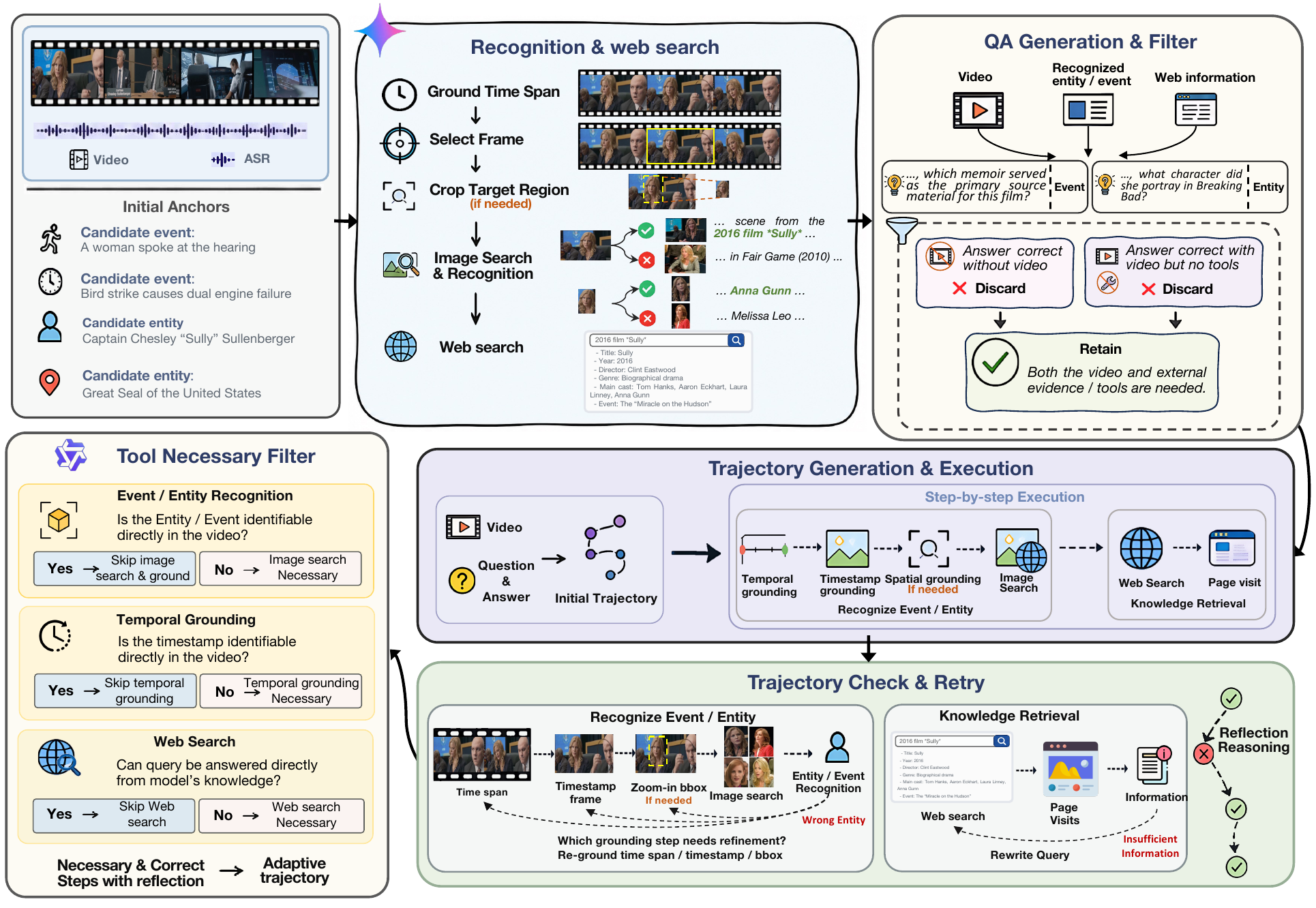}
    }{
        \fbox{\rule{0pt}{2in}\rule{0.98\linewidth}{0pt}}
    }
    \vspace{-10pt}
    \caption{Overview of the data construction pipeline, including QA and trajectory generation, trajectory refinement, and tool-necessary filtering.
    }
    \label{fig:data_collection}
    \vspace{-14pt}
\end{figure}

\vspace{-4pt}
\section{Data Collection}
\label{section:data_collection}

To enable the adaptive behaviors described above, we construct training data that reflects task-specific tool requirements and model-specific tool necessity. As illustrated in Fig.~\ref{fig:data_collection}, our pipeline consists of two stages: \emph{QA generation} and \emph{trajectory generation}, where we construct questions requiring both video and external knowledge, and build verified trajectories with unnecessary tool interactions removed.

\subsection{QA Generation}
\label{sec:qa_generation}

We first perform \emph{video selection and anchor discovery} to identify retrieval-relevant events and entities from diverse videos. We then conduct \emph{video-to-Web evidence acquisition}, progressively grounding the target content and retrieving related external knowledge. Finally, through \emph{QA generation and filtering}, we construct questions that jointly require video information and external evidence, and remove those that can be answered without either source.

\noindent\textbf{Video Selection and Anchor Discovery.} We primarily collect videos from YouTube for the news, industry, culture, entertainment, sports, and science domains. The scene-understanding subset is instead primarily sourced from EgoExo4D \citep{grauman2024ego}, EgoLife \citep{yang2025egolife}, CityWalker \citep{liu2025citywalker}, ARKitScenes \citep{baruch2021arkitscenes}, ScanNet++ \citep{yeshwanth2023scannet++}, and Sekai \citep{li2026sekai}, covering diverse indoor, outdoor, egocentric, and urban environments.
The resulting collection covers seven semantic domains: news, industry, culture, entertainment, sports, science, and scene understanding.

Gemini-3.5-Flash analyzes each video with its ASR transcript to identify fine-grained and retrieval-relevant visual anchors. At this stage, the model does not need to determine the exact identity of each entity or event. Instead, it records textual descriptions of potentially useful visual content for subsequent retrieval and question construction. These descriptions include both recognizable entities or events with clear retrieval value and unfamiliar visual targets whose identities can be further resolved through search.

\noindent\textbf{Video-to-Web Evidence Acquisition.}
To collect fine-grained external knowledge, we first extract sufficient information from the video to identify the relevant event or entity, and then use the extracted information to conduct further image retrieval or web retrieval. This process consists of two successive stages: video-based information extraction and external evidence retrieval.

For \emph{video-based information extraction}, we progressively identify the target through multi-step grounding and retrieval. For events or actions, we perform temporal grounding to locate the relevant segment and timestamp grounding to select informative frames.
The selected frames are then used for image search to retrieve visually related results, which are combined with complementary textual information from ASR.
For entities, we perform spatial grounding to isolate the target region and use the resulting crop for image search to obtain possible identities. OCR and ASR further provide visible or spoken information.
We combine these results until sufficiently specific information about the event or entity is obtained for subsequent retrieval.

Depending on the identified event or entity, we then perform \emph{external evidence retrieval} by using web search or image search to retrieve relevant evidence from authoritative or task-specific sources, and organize the retrieved facts into a traceable evidence path.

\noindent\textbf{QA Generation and Filtering.}
Based on the verified video information and source-backed external evidence, we construct question--answer pairs that require reasoning over both sources. Each question combines video-grounded information with a temporal or timestamp-related condition, such as an event occurring within a particular interval or an entity appearing at a specific moment, while the answer is derived from the corresponding external evidence.

To ensure that the resulting questions genuinely depend on both the video and external information, we further filter the generated QAs using Qwen3-VL-235B-A22B-Instruct under two restricted settings: (1) answering without access to the video, and (2) answering with the video but without using external tools. We remove a question if it can be answered correctly under either setting.

\subsection{Trajectory Generation}
\label{sec:trajectory_generation}

For each retained QA pair, we first perform \emph{trajectory generation and execution}, constructing task-specific tool-use trajectories, refining their logical dependencies and parameters, and executing them step by step. We then conduct \emph{trajectory check and retry} to verify intermediate results, retry unreliable steps when necessary, and preserve the resulting correction process as adaptive reflection data.
Finally, through \emph{model-conditioned tool necessity filtering}, we further remove tool calls or tool chains that are unnecessary for the target model based on its own video understanding capability and internal knowledge.

\noindent\textbf{Trajectory Generation and Execution.}
For each retained QA pair, we first organize the information acquired in the previous stage, including video grounding results, event or entity identification, and the corresponding external evidence. Different questions follow different information acquisition processes through task-specific prompts; for example, event-centric questions are explicitly instructed not to invoke spatial grounding. Based on these processes, we further organize the grounding and retrieval operations, together with their intermediate results, into corresponding tool-use trajectories that are aligned with the specific information requirements of each question.

We refine each initial trajectory according to the question type, removing tool calls that are irrelevant to its information requirements. For example, spatial grounding is removed for event-centric questions when no fine-grained spatial localization is needed. We then check the logical dependencies and tool parameters of the remaining trajectory, ensuring that each tool is supported by the outputs of its prerequisite steps. For instance, image search over a target entity should be preceded by the corresponding grounding operation, with its input taken from the resulting crop. We further verify that each tool call correctly refers to the intended intermediate results and revise invalid dependencies or parameters accordingly. We then invoke the tools to execute the refined trajectory step by step according to the specified tool calls and parameters.

\noindent\textbf{Trajectory Check and Retry.}
We then execute the refined trajectory and verify whether each tool produces the expected information.
For image search, an incorrect result often originates from low-quality grounding, such as selecting a frame where a person is shown only from the side or cropping an uninformative region. In such cases, we trace back to the corresponding grounding step, select a better frame or region, and repeat the image search until the expected target appears in the retrieved results or the maximum number of attempts is reached. For web search, when the retrieved results do not contain the expected information, we reformulate the search query and repeat the retrieval. This iterative execution and verification process continues until the trajectory can reliably recover the expected evidence or the maximum number of attempts is reached.

We retain these failed attempts, failure diagnoses, and subsequent corrections in the trajectory, producing natural reflection and recovery steps that teach the model to revise unreliable intermediate results. By introducing reflection only when unreliable intermediate results occur, while leaving reliable steps unchanged, the constructed trajectories teach the agent to trigger reflection on demand and perform targeted correction rather than reflecting at every reasoning step.

\noindent\textbf{Model-conditioned Tool Necessity Filtering.}
We further evaluate the necessity of tool calls in each successfully executed trajectory.
Given a tool or a consecutive tool chain $\mathcal{S}$, we denote the information already known to the agent before executing $\mathcal{S}$ as $x_{\mathcal{S}}$, and the target result obtained after executing $\mathcal{S}$ as $y_{\mathcal{S}}$.
Based on the parameters of $\mathcal{S}$, such as the grounding description or retrieval target, we construct a verification query $q_{\mathcal{S}}$ that asks for the target result $y_{\mathcal{S}}$. We use the base model that will initialize the agent for subsequent training, denoted by $\mathcal{M}_{\phi}$, to perform necessity filtering. In our experiments, this model is Qwen3.5-9B, the base model of AdaVDR-9B. Tool necessity is therefore evaluated against the video understanding capability and internal knowledge of the model to be trained. The base model then directly obtains the target result from $x_{\mathcal{S}}$ without executing $\mathcal{S}$:
\begin{equation}
    \begin{aligned}
        \hat{y}_{\mathcal{S}}
        =
        \mathcal{M}_{\phi}
        \left(
        x_{\mathcal{S}},
        q_{\mathcal{S}}
        \right).
    \end{aligned}
\end{equation}

The necessity of $\mathcal{S}$ is defined as
\begin{equation}
    \begin{aligned}
        N(\mathcal{S})
        =
        \mathbf{1}
        \left[
            \hat{y}_{\mathcal{S}}
            \neq
            y_{\mathcal{S}}
            \right].
    \end{aligned}
\end{equation}

If $\hat{y}_{\mathcal{S}}$ matches $y_{\mathcal{S}}$, the target result can be obtained without executing $\mathcal{S}$, indicating that the tool chain is unnecessary and can be removed. Otherwise, $\mathcal{S}$ is retained as necessary.

We apply this evaluation to the longest candidate tool chain and then progressively examine its individual tools.
For example, consider the chain
\texttt{Temporal Grounding} $\rightarrow$ \texttt{Timestamp Grounding} $\rightarrow$ \texttt{Spatial Grounding} $\rightarrow$ \texttt{Image Search}, which is used to identify ``the person wearing blue.''
If the base model can directly identify the person from the original video, the entire tool chain is removed.
Then, we further evaluate the necessity of the tools within the chain. For example, to evaluate \texttt{Temporal Grounding}, we bypass its localized temporal interval and directly provide the original video together with a query asking the model to locate a usable frame containing ``the person wearing blue.''
If the base model can directly locate such a frame from the full video, \texttt{Temporal Grounding} is unnecessary and removed.
We apply this principle to the remaining grounding and retrieval tools, progressively removing tool calls.

\subsection{Training Data}
\label{sec:training_data}

Following the above pipeline, we automatically construct SFT data with tool-use trajectories. Different question types correspond to different tool-use trajectories, while model capability filtering retains only necessary tool calls, enabling the agent to acquire adaptive capabilities through SFT. For RL data, we construct QA pairs following the procedure in Sec.~\ref{sec:qa_generation}.

\begingroup
\setlength{\intextsep}{6pt}
\begin{table}[H]
    \centering
    \begin{minipage}[c]{0.35\textwidth}
        \centering
        \includegraphics[width=1\linewidth]{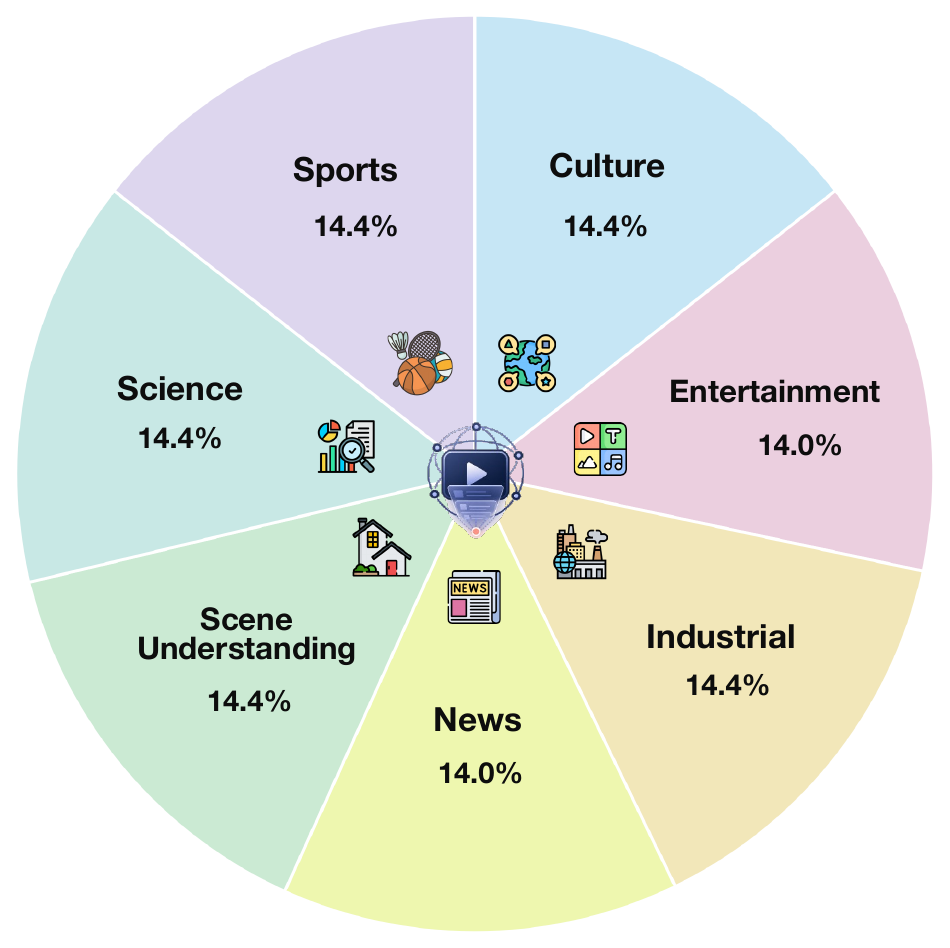}
        \captionof{figure}{Distribution of the 250 VDR-EE questions across seven semantic domains.}
        \label{fig:benchmark_domains}
    \end{minipage}\hfill
    \begin{minipage}[c]{0.63\textwidth}
        \centering
        \captionof{table}{Statistics and video-centric annotations of VDR-EE.}
        \label{tab:dataset_statistics}
        \vspace{3pt}
        {\small\setlength{\tabcolsep}{1.4pt}
            \renewcommand{\arraystretch}{1.05}
            \begin{tabular}{@{}>{\raggedright\arraybackslash}p{0.24\linewidth}@{\hspace{5pt}}>{\raggedright\arraybackslash}p{0.15\linewidth}@{\hspace{5pt}}>{\raggedright\arraybackslash}p{0.24\linewidth}@{\hspace{4pt}}r@{\hspace{4pt}}r@{}}
                \toprule
                \textbf{Dimension}    & \textbf{Category} & \textbf{Definition}       & \textbf{Count} & \textbf{Proportion} \\
                \midrule
                \rowcolor{gray!8}\multicolumn{5}{l}{\textit{Question Type (250 instances)}}                                  \\
                \mbox{Question Type}  & Entity            & Entity-oriented questions & 153            & 61.20\%             \\
                                      & Event             & Event-oriented questions  & 97             & 38.80\%             \\
                \rowcolor{gray!8}\multicolumn{5}{l}{\textit{Entity Multiplicity (153 entity instances)}}                     \\
                \mbox{Entity Type}    & Single            & \mbox{One entity}         & 76             & 49.67\%             \\
                                      & Multi             & \mbox{Multiple entities}  & 77             & 50.33\%             \\
                \rowcolor{gray!8}\multicolumn{5}{l}{\textit{Grounded-Event Duration (97 event instances)}}                   \\
                \mbox{Event Duration} & Short             & $\leq 30$ s               & 35             & 36.08\%             \\
                                      & Medium            & $30$--$120$ s             & 34             & 35.05\%             \\
                                      & Long              & $>120$ s                  & 28             & 28.87\%             \\
                \rowcolor{gray!8}\multicolumn{5}{l}{\textit{Full-Video Duration (250 instances)}}                            \\
                \mbox{Video Duration} & Short             & $\leq 5$ min              & 77             & 30.80\%             \\
                                      & Medium            & $5<t<10$ min              & 80             & 32.00\%             \\
                                      & Long              & $\geq 10$ min             & 93             & 37.20\%             \\
                \bottomrule
            \end{tabular}%
        }
    \end{minipage}
\end{table}
\endgroup

\subsection{Benchmark}
\label{sec:benchmark}
\vspace{-2pt}

By the QA generation process, we introduce \textbf{VDR-EE}, a manually verified Video Deep Research benchmark.  Each QA pair is manually verified to ensure clear identification of the related entity or event in the video, dependence on both video and external evidence, question clarity, and answer support from retrieved evidence. We retain 250 manually verified questions.

\noindent\textbf{\emph{Semantic domain.}} Each retained question is manually assigned to one of seven domains: culture, entertainment, industry, news, scene understanding, science, and sports. The resulting distribution across these seven domains is shown in Fig.~\ref{fig:benchmark_domains}.

\begin{figure}[htbp]
    \centering
    \includegraphics[width=0.95\linewidth]{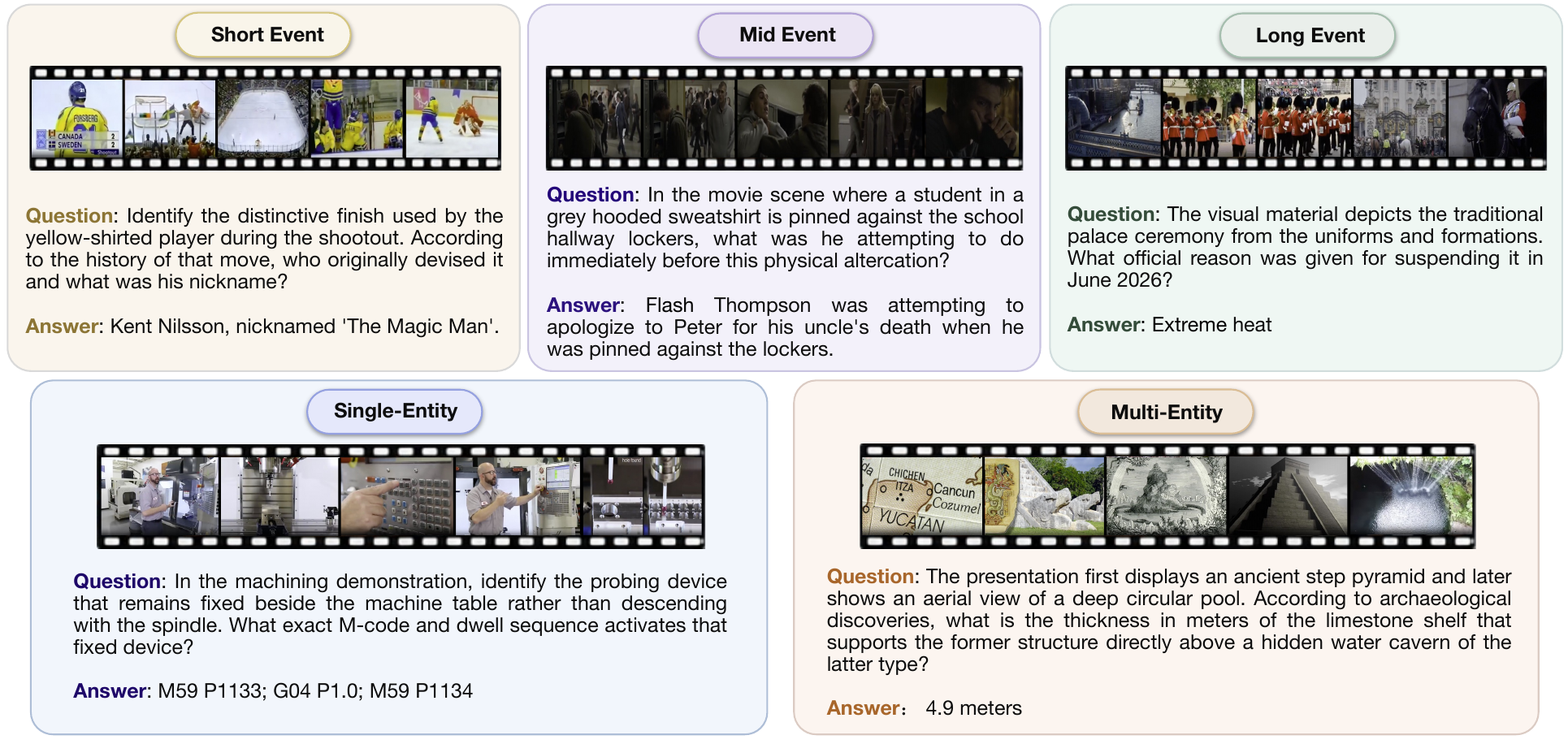}
    \caption{Examples of event-oriented and entity-oriented questions in VDR-EE.}
    \label{fig:benchmark_cases}
\end{figure}

\noindent\textbf{\emph{Video-centric categories.}} VDR-EE contains entity-centric and event-centric questions. 
We further divide entity questions into single-entity questions and multi-entity questions. Event questions are grouped by the duration of the grounded target event: short ($\leq 30$ s), medium ($30$--$120$ s), and long ($>120$ s). All questions are also grouped by full-video duration into short ($\leq 5$ min), medium ($5<t<10$ min), and long ($\geq 10$ min) videos. Table~\ref{tab:dataset_statistics} summarizes these annotations, including 153 entity and 97 event questions. Representative questions are shown in Fig.~\ref{fig:benchmark_cases}.

These annotation dimensions capture different localization demands: entity questions may require identifying multiple visual targets, while event questions may require locating an interval before external retrieval.

\vspace{-4pt}
\section{Training}
\label{section:training}
\vspace{-4pt}

\begin{figure}[hbtp!]
    \centering
    \includegraphics[width=\linewidth]{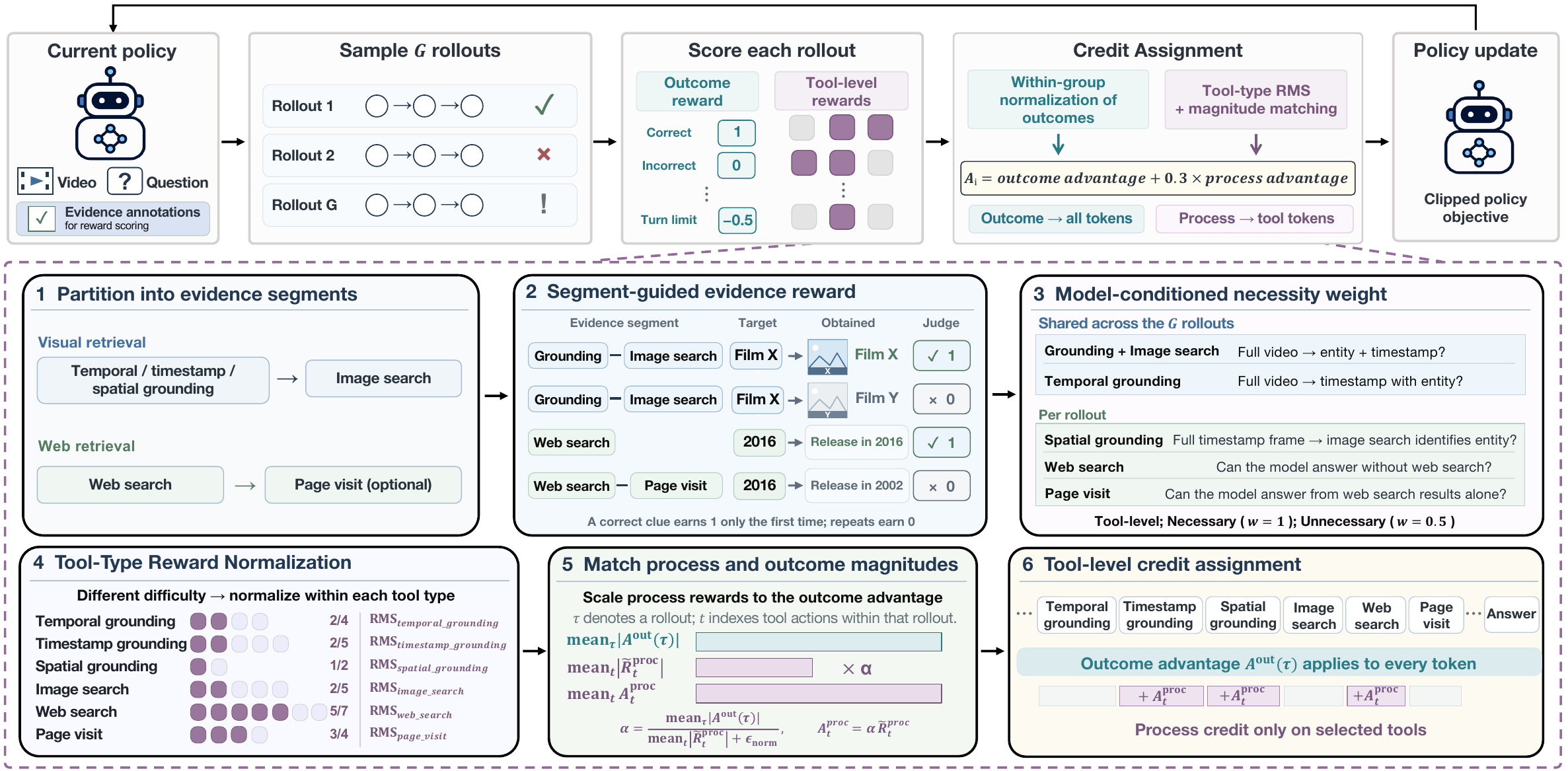}
    \vspace{-2pt}
    \caption{Overview of NASA-GRPO, a necessity-aware segment-guided credit assignment RL.}
    \vspace{-10pt}
    \label{fig:nasa_grpo}
\end{figure}

\subsection{Cold-Start Supervised Fine-Tuning}

We perform supervised fine-tuning (SFT) on 2.9K verified trajectories from Sec.~\ref{sec:trajectory_generation} to initialize adaptive tool use and reflection.
Given $M$ trajectories, where $x^{(n)}$ denotes the interaction context and $y^{(n)}=\{y^{(n)}_1,\ldots,y^{(n)}_{N_n}\}$ the target assistant output, we optimize:
    \vspace{-4pt}
\begin{equation}
\small
    \mathcal{L}_{\mathrm{SFT}}(\theta)
    = -\frac{1}{M}\sum_{n=1}^{M}
    \sum_{i=1}^{N_n}
    \log \pi_{\theta}\!\left(
    y^{(n)}_i \mid x^{(n)}, y^{(n)}_{<i}
    \right).
    \label{eq:sft}
\end{equation}
The loss is applied only to assistant-generated tokens, while video inputs, user messages, and tool observations serve as conditioning context. The resulting checkpoint serves as the initial RL policy.

\vspace{-2pt}
\subsection{NASA-GRPO}
\vspace{-2pt}


In video deep research, a rollout typically involves multiple tool calls, some of which may fail or be redundant. Relying solely on the outcome reward  provides sparse  supervision, making it difficult to learn tool-use policies. Moreover, evidence often requires a sequence of consecutive tool calls to obtain. We propose NASA-GRPO, a necessity-aware segment-guided credit assignment method for GRPO, which assigns tool-level rewards according to their importance within each evidence-acquisition segment. 
As shown in Fig.~\ref{fig:nasa_grpo}, NASA-GRPO partitions rollouts into evidence-acquisition segments, scores each segment's evidence contribution and each tool's necessity, normalizes rewards by tool type, and combines them with the outcome reward.

\paragraph{Outcome Reward.} Given a training sample $(V,q,a^*)$, the outcome reward of a rollout $\tau$ is
\begin{equation}
\small
    R^{\mathrm{out}}(\tau)
    =
    \begin{cases}
        1,    & \text{if the final answer is correct},                      \\
        0,    & \text{if the final answer is incorrect},                    \\
        -0.5, & \text{if the rollout is invalid or reaches the turn limit}.
    \end{cases}
    \label{eq:outcome_reward}
\end{equation}
GRPO normalizes the outcome reward within a sampled rollout group $\mathcal{G}$, which is computed as
\begin{equation}
\small
    A^{\mathrm{out}}(\tau)
    =
    \frac{R^{\mathrm{out}}(\tau)-\mu_{\mathcal{G}}}
    {\sigma_{\mathcal{G}}+\varepsilon_{\mathrm{norm}}},
    \label{eq}
\end{equation}
where $\mu_{\mathcal{G}}$ and $\sigma_{\mathcal{G}}$ denote the group mean and standard deviation.

\paragraph{Segment-Guided Evidence Reward.} Given $(V,q,a^*)$, our data construction provides target clues $\mathcal{C}^*$, including video entities and external facts required to answer the question.
We partition rollout $\tau=(a_1,o_1,\ldots,a_T,o_T)$ into segments $\{\mathcal{S}_m\}_{m=1}^{M}$, each of which corresponds to one external information-acquisition process, including both the tool actions that prepare the retrieval input and the retrieval actions that obtain the target information. Each segment targets one clue in $\mathcal{C}^*$. For visual retrieval, a segment includes grounding and image-search tools. For Web retrieval, a segment includes a Web-search tool and may include page-visit tool.
Let $\widehat{c}_m$ denote the clue obtained by $\mathcal{S}_m$. We assign evidence reward only when the clue is both correct and has not been obtained,
\begin{equation}
\small
    S^{\mathrm{evid}}_m
    =
    \mathbb{I}\left[J(\widehat{c}_m,c_m^*)=1\right]
    \cdot
    \mathbb{I}\left[c_m^*\notin\mathcal{C}^{\mathrm{credit}}_{<m}\right],
    \label{eq:evidence_credit}
\end{equation}
where $\mathcal{C}^{\mathrm{credit}}_{<m}$ contains the target clues credited in earlier segments. The first term measures clue correctness, while the second measures whether the clue has been previously obtained.




\paragraph{Model-Conditioned Necessity Weight.}
Evidence contribution does not imply tool necessity. A tool may return a correct clue even when the target model can obtain the same information without using it. Such a tool call is not necessary. We assign it less credit so that the policy is not encouraged to make unnecessary tool calls.

We evaluate necessity by bypassing the tool and asking the current policy to recover the same target information. At each RL iteration, we use the policy snapshot $\pi_{\theta_{\mathrm{old}}}$ that generates the rollout group. This snapshot is held fixed during reward evaluation and refreshed at the next iteration, so necessity judgments track the evolving capabilities of the policy. Some evaluations use information shared by all rollouts in the same GRPO group. We first provide the full video and ask the model to identify the target entity and its timestamp. If the model identifies the entity correctly, \texttt{image\_search} and its preceding grounding actions are unnecessary. Otherwise, the visual retrieval chain is retained. We then ask the model to locate a timestamp that contains the target entity directly from the full video. If the returned timestamp contains the correct entity, the preceding \texttt{temporal\_grounding} action is unnecessary; otherwise, temporal grounding is necessary. These evaluations are performed once for the rollout group because its rollouts share the same video, question, and target entities.

Other evaluations depend on the intermediate result or information target of a specific rollout. We therefore evaluate them separately for each rollout. For \texttt{spatial\_grounding}, we use the full frame selected by that rollout and perform image search without the predicted crop. If the full frame is sufficient to identify the target entity, spatial grounding is not necessary. Otherwise, the crop is necessary. We perform this evaluation only for rollouts that invoke spatial grounding. For \texttt{web\_search}, we ask the model to answer the knowledge question targeted by the current rollout without using external retrieval.

An action receives full reward when the model cannot recover its target information without the action. Otherwise, the action retains half reward.
Mathematically, for an action $a_t$ in segment $s(t)$, the  necessity weight is computed as
\begin{equation}
\small
    w_t^{\mathrm{nec}}
    =
    \begin{cases}
        1,   & \text{if } \widehat{y}_{a_t}^{-} \neq y_{a_t}, \\
        0.5, & \text{if } \widehat{y}_{a_t}^{-} = y_{a_t},
    \end{cases}
    \label{eq:necessity_weight}
\end{equation}
where $y_{a_t}$ denotes its target information,  $h_t^{-}$ denotes the input used after bypassing the action, and  $ \widehat{y}_{a_t}^{-}
    =
    \pi_{\theta_{\mathrm{old}}}(h_t^{-},q_{a_t}^{\mathrm{verify}})$ denotes the output of the current policy snapshot without $a_t$. Combining the segment-guided evidence credit, the credit of action $a_t$ is computed as
\begin{equation}
\small
    R^{\mathrm{proc}}_t
    =
    S^{\mathrm{evid}}_{s(t)}
    \cdot
    w_t^{\mathrm{nec}}.
    \label{eq:process_reward}
\end{equation}

\paragraph{Tool-Type Reward Normalization.}
Different tool types receive positive rewards at different rates due to differences in task difficulty. We therefore normalize process rewards within each tool type to prevent easier tools from dominating the learning signal. Let $k(t)$ denote the type of action $a_t$, and let $\mathcal{B}_k$ contain all actions of type $k$ in the rollout batch. The process reward is normalized as
\begin{equation}
    \small
    \widetilde{R}^{\mathrm{proc}}_t
    =
    \frac{R^{\mathrm{proc}}_t}
    {\sqrt{
            \frac{1}{|\mathcal{B}_{k(t)}|}
            \sum_{j\in\mathcal{B}_{k(t)}}
            \left(R^{\mathrm{proc}}_j\right)^2
            +\varepsilon_{\mathrm{norm}}
        }}.
    \label{eq:tool_normalization}
\end{equation}
We further match the average magnitude of the process signal to that of the outcome advantage:
\begin{equation}
    \small
    \alpha
    =
    \frac{\operatorname{mean}_{\tau}|A^{\mathrm{out}}(\tau)|}
    {\operatorname{mean}_{t}|\widetilde{R}^{\mathrm{proc}}_t|+\varepsilon_{\mathrm{norm}}},
    \qquad
    A^{\mathrm{proc}}_t
    =
    \alpha\widetilde{R}^{\mathrm{proc}}_t.
    \label{eq:process_advantage}
\end{equation}

\paragraph{Tool-Level Credit Assignment.}
The outcome advantage is applied to all tokens. The tool-level advantage is applied only to the tokens of the credited tool action. Let $m_{t,i}=1$ if token $i$ belongs to action $a_t$, and $m_{t,i}=0$ otherwise. The final token reward is
\begin{equation}
\small
    A_i
    =
    A^{\mathrm{out}}(\tau)
    +
    \lambda
    \sum_t m_{t,i}A^{\mathrm{proc}}_t,
    \qquad \lambda=0.3.
    \label{eq:final_advantage}
\end{equation}

\vspace{-8pt}
\section{Experiment}
\vspace{-4pt}

\subsection{Implementation Details}
\vspace{-2pt}
We evaluate AdaVDR-9B, initialized from Qwen3.5-9B~\citep{team2026qwen3}, on VDR-EE and VideoDR~\citep{liu2026watching}. Training details for AdaVDR-9B are provided in Appendix~\ref{app:training_details} and Appendix~\ref{app:training_dynamics}.
Baselines include Gemini-3.1-Pro and Gemini-3-Flash~\citep{team2023gemini}, GPT-5.4~\citep{singh2025openai}, Qwen3.5-35B-A3B and Qwen3.5-9B~\citep{team2026qwen3}, and Qwen3-VL-32B-Instruct and Qwen3-VL-8B-Instruct~\citep{bai2025qwen3}. We also include VideoSearcher-8B~\citep{gao2026videosearcher}, VideoRover-8B~\citep{liu2026thinking}, and Video-DeepResearch-30B-A3B~\citep{fang2026video}.
Direct answers from the video alone; Workflow separates video-cue extraction from Web retrieval; Agentic lets a multimodal agent coordinate video understanding, tool use, and evidence integration. We report AdaVDR-9B under Agentic and use GPT-5.4 to judge answer accuracy.

\begingroup
\begin{table}[htbp]
    \centering
    \caption{Accuracy (\%) on benchmkarks.
        VDR-EE uses the \textbf{Direct} and \textbf{Agentic} settings, while VideoDR uses the \textbf{Workflow} and \textbf{Agentic} settings. AdaVDR results are averaged over three runs.
    }
    \vspace{-2pt}
    \label{tab:our_bench_eval1}
    \resizebox{\linewidth}{!}{%
        \begin{tabular}{lr|rrrrrrrr|r}
            \toprule
            \textbf{Model}                                   & \textbf{VideoDR}                       & \multicolumn{8}{c|}{\textbf{VDR-EE}} & \textbf{Avg.}                                                                                                                                                                                                                                                                    \\
            \cmidrule(lr){2-2}\cmidrule(lr){3-10}\cmidrule(lr){11-11}
                                                             & \textbf{Avg.}                          & \multicolumn{3}{c}{\textbf{Entity}}  & \multicolumn{4}{c}{\textbf{Event}} & \textbf{Avg.}                                                                                                                                                                                                              &  \\
            \cmidrule(lr){3-5}\cmidrule(lr){6-9}
                                                             &                                        & \textbf{Single}                      & \textbf{Multiple}                  & \textbf{Overall}                  & \textbf{Short}                     & \textbf{Medium}                    & \textbf{Long}                      & \textbf{Overall}                   &                                    &  \\
            \midrule
                                                             & \multicolumn{1}{c|}{\textit{Workflow}} & \multicolumn{8}{c|}{\textit{Direct}} &                                                                                                                                                                                                                                                                    \\
            \midrule
            \rowcolor{gray!10}\multicolumn{11}{l}{\textit{Proprietary}}                                                                                                                                                                                                                                                                                                                                        \\
            Gemini-3.1-Pro~\citep{team2023gemini}            & --                                     & 56.58                                & 28.57                              & 42.48                             & 54.29                              & 55.88                              & 50.00                              & 53.61                              & 46.80                              & -- \\
            Gemini-3-Flash~\citep{team2023gemini}            & 59.00                                  & 42.11                                & 22.08                              & 32.03                             & 51.43                              & 61.76                              & 46.43                              & 53.61                              & 40.40                              & 49.70 \\
            GPT-5.4~\citep{singh2025openai}                  & 51.00                                  & 35.53                                & 23.38                              & 29.41                             & 28.57                              & 41.18                              & 39.29                              & 36.08                              & 32.00                              & 41.50 \\
            \rowcolor{blue!6}\multicolumn{11}{l}{\textit{Open-source}}                                                                                                                                                                                                                                                                                                                                         \\
            Qwen3-VL-8B-Instruct~\citep{bai2025qwen3}        & 28.00                                  & 21.05                                & 5.19                               & 13.07                             & 8.57                               & 14.71                              & 10.71                              & 11.34                              & 12.40                              & 20.20 \\
            Qwen3-VL-32B-Instruct~\citep{bai2025qwen3}       & 36.00                                  & 30.26                                & 9.09                               & 19.61                             & 8.57                               & 14.71                              & 14.29                              & 12.37                              & 16.80                              & 26.40 \\
            Qwen3.5-9B~\citep{team2026qwen3}                 & 35.00                                  & 23.68                                & 11.69                              & 17.65                             & 5.71                               & 2.94                               & 7.14                               & 5.15                               & 12.80                              & 23.90 \\
            Qwen3.5-35B-A3B~\citep{team2026qwen3}            & 39.00                                  & 32.89                                & 9.09                               & 20.92                             & 11.43                              & 5.88                               & 7.14                               & 8.25                               & 16.00                              & 27.50 \\
            \midrule
                                                             & \multicolumn{1}{c|}{\textit{Agentic}}  & \multicolumn{8}{c|}{\textit{Agentic}} &                                                                                                                                                                                                                                                                   \\
            \midrule
            \rowcolor{gray!10}\multicolumn{11}{l}{\textit{Proprietary}}                                                                                                                                                                                                                                                                                                                                        \\
            Gemini-3.1-Pro~\citep{team2023gemini}            & --                                     & 73.68                                & 50.65                              & 62.09                             & 71.43                              & 64.71                              & 67.86                              & 68.04                              & 64.40                              & -- \\
            Gemini-3-Flash~\citep{team2023gemini}            & --                                     & 64.47                                & 40.26                              & 52.29                             & 68.57                              & 82.35                              & 60.71                              & 71.13                              & 59.60                              & -- \\
            GPT-5.4~\citep{singh2025openai}                  & --                                     & 59.21                                & 40.26                              & 49.67                             & 54.29                              & 61.76                              & 57.14                              & 57.73                              & 52.80                              & -- \\
            \rowcolor{blue!6}\multicolumn{11}{l}{\textit{Open-source}}                                                                                                                                                                                                                                                                                                                                         \\
            Qwen3-VL-8B-Instruct~\citep{bai2025qwen3}        & 30.00                                  & 42.11                                & 22.08                              & 32.03                             & 22.86                              & 26.47                              & 17.86                              & 22.68                              & 28.40                              & 29.20 \\
            Qwen3-VL-32B-Instruct~\citep{bai2025qwen3}       & 38.00                                  & 60.53                                & 32.47                              & 46.41                             & 22.86                              & 32.35                              & 17.86                              & 24.74                              & 38.00                              & 38.00 \\
            Qwen3.5-9B~\citep{team2026qwen3}                 & 37.00                                  & 56.58                                & 28.57                              & 42.48                             & 34.29                              & 32.35                              & 42.86                              & 36.08                              & 40.00                              & 38.50 \\
            Qwen3.5-35B-A3B~\citep{team2026qwen3}            & 42.00                                  & 56.58                                & 38.96                              & 47.71                             & 42.86                              & 52.94                              & 35.71                              & 44.33                              & 46.40                              & 44.20 \\
            VideoSearcher-8B~\citep{gao2026videosearcher}    & 53.00                                  & {53.95}                              & {32.47}                            & {43.14}                           & {54.29}                            & {55.88}                            & {53.57}                            & {54.64}                            & {47.60}                            & 50.30 \\
            VideoRover-8B~\citep{liu2026thinking}            & 56.00                                  & --                                   & --                                 & --                                & --                                 & --                                 & --                                 & --                                 & --                                 & -- \\
            Video-DeepResearch-30B-A3B~\citep{fang2026video} & 62.00                                  & {61.84}                              & {36.36}                            & {49.02}                           & {62.86}                            & {64.71}                            & {60.71}                            & {62.89}                            & {54.40}                            & 58.20 \\
            \rowcolor{green!5}\textbf{AdaVDR-9B-SFT (Qwen3.5-9B)}
                & 52.00 & 60.53 & 33.77 & 47.06 & 48.57 & 47.06 & 53.57 & 49.48 & 48.00 & 50.00 \\
            \textcolor{green!50!black}{$\Delta$ vs. Qwen3.5-9B}
                & \textcolor{green!50!black}{+15.00} & \textcolor{green!50!black}{+3.95} & \textcolor{green!50!black}{+5.20} & \textcolor{green!50!black}{+4.58} & \textcolor{green!50!black}{+14.28} & \textcolor{green!50!black}{+14.71} & \textcolor{green!50!black}{+10.71} & \textcolor{green!50!black}{+13.40} & \textcolor{green!50!black}{+8.00} & \textcolor{green!50!black}{+11.50} \\
            \rowcolor{green!10}\textbf{AdaVDR-9B (Qwen3.5-9B)}
                                                             & \textbf{66.00}                         & \textbf{63.16}                       & \textbf{37.66}                     & \textbf{50.33}                    & \textbf{62.86}                     & \textbf{64.71}                     & \textbf{60.71}                     & \textbf{62.89}                     & \textbf{55.20}                     & \textbf{60.60} \\
            \textcolor{green!50!black}{$\Delta$ vs. Qwen3.5-9B}
                                                             & \textcolor{green!50!black}{+29.00}     & \textcolor{green!50!black}{+6.58}    & \textcolor{green!50!black}{+9.09}  & \textcolor{green!50!black}{+7.85} & \textcolor{green!50!black}{+28.57} & \textcolor{green!50!black}{+32.36} & \textcolor{green!50!black}{+17.85} & \textcolor{green!50!black}{+26.81} & \textcolor{green!50!black}{+15.20} & \textcolor{green!50!black}{+22.10} \\
            \bottomrule
        \end{tabular}%
    }
\end{table}
\endgroup

\subsection{Main Results}

We evaluate the models on VDR-EE under the Direct and Agentic settings and on VideoDR under the Workflow and Agentic settings.
Results are shown in Table~\ref{tab:our_bench_eval1}.
Additional results on VDR-EE and VideoDR, as well as evaluations on image deep research and long-video understanding, are provided in Appendix~\ref{app:more_results}.
Qualitative examples are provided in Sec.~\ref{sec:qualitative_analysis}.

\noindent SFT improves the Qwen3.5-9B base model on both benchmarks. As shown in Table~\ref{tab:our_bench_eval1}, AdaVDR-9B-SFT raises accuracy from 37.00\% to 52.00\% on VideoDR and from 40.00\% to 48.00\% on VDR-EE, corresponding to gains of 15.00 and 8.00 percentage points, respectively. Subsequent NASA-GRPO training further improves accuracy to 66.00\% and 55.20\%, adding 14.00 and 7.20 percentage points over SFT.

\noindent AdaVDR consistently improves performance on both benchmarks. AdaVDR outperforms the base model Qwen3.5-9B on both VDR-EE and VideoDR, improving average accuracy by 15.20 and 29.00 percentage points, respectively. The improvements are consistent across entity- and event-centric questions, all seven domains on VDR-EE, and different domains and difficulty levels on VideoDR (Tables~\ref{tab:our_bench_eval2}, \ref{tab:video_deepresearch_results}, and~\ref{tab:difficulty_results}). These results demonstrate the superiority of the proposed method. Agentic interaction consistently improves performance. For all baselines with paired results, the Agentic setting outperforms the Direct setting on VDR-EE and the Workflow setting on VideoDR (Table~\ref{tab:our_bench_eval1}). These results show the benefit of allowing the model to autonomously coordinate video grounding and external retrieval rather than relying on direct answering or a predefined workflow.


\vspace{-4pt}
\subsection{Tool-Use Analysis}
\vspace{-2pt}

\begin{table}[H]
    \centering
    \caption{Average tool calls and turns per question across training stages on VideoDR and VDR-EE.}
    \vspace{-4pt}
    \label{tab:tool_analysis}
    \renewcommand{\arraystretch}{1.12}
    {\setlength{\tabcolsep}{4pt}\resizebox{0.85\linewidth}{!}{%
            \begin{tabular}{lrrrrrrr}
                \toprule
                \textbf{Model} & \multicolumn{3}{c}{\textbf{Video Grounding}} & \multicolumn{3}{c}{\textbf{External Retrieval}} & \textbf{Trajectory}                                                                                           \\
                \cmidrule(lr){2-4}\cmidrule(lr){5-7}\cmidrule(lr){8-8}
                               & \textbf{Temporal}                            & \textbf{Timestamp}                              & \textbf{Spatial}    & \textbf{Image Search} & \textbf{Web Search} & \textbf{Visit Page} & \textbf{Avg. Turns} \\
                \midrule
                \rowcolor{gray!8}\multicolumn{8}{l}{\textit{VideoDR}}                                                                                                                                                                           \\
                Qwen3.5-9B     & 0.46                                         & 0.24                                            & 0.07                & 0.12                  & 2.56                & 0.67                & 5.12                \\
                AdaVDR-9B-SFT  & 1.55                                         & 1.73                                            & 0.44                & 0.93                  & 4.07                & 0.46                & 10.18               \\
                AdaVDR-9B-RL
                               & 1.09                                         & 1.14                                            & 0.02                & 1.09                  & 4.81                & 2.52                & 11.67               \\
                \midrule
                \rowcolor{gray!8}\multicolumn{8}{l}{\textit{VDR-EE}}                                                                                                                                                                            \\
                Qwen3.5-9B     & 0.99                                         & 0.79                                            & 0.12                & 0.40                  & 2.12                & 0.28                & 5.70                \\
                AdaVDR-9B-SFT  & 1.28                                         & 1.16                                            & 0.78                & 0.99                  & 5.05                & 0.55                & 10.81               \\
                AdaVDR-9B-RL
                               & 1.04                                         & 1.11                                            & 0.06                & 1.09                  & 5.54                & 3.42                & 13.26               \\
                \bottomrule
            \end{tabular}}}
    \vspace{-6pt}
    \renewcommand{\arraystretch}{1.0}
\end{table}

Table~\ref{tab:tool_analysis} shows that SFT makes more tool calls than the base model on both benchmarks. Compared with SFT, NASA-GRPO makes fewer grounding calls but more image searches, web searches, and page visits. On the one hand, SFT may hallucinate citations when retrieval is inconclusive, whereas NASA-GRPO rewards obtaining supporting evidence. On the other hand, correct grounding does not guarantee a useful image-search result, so NASA-GRPO may favor more informative frames and crops. Accuracy improves by 14.0 and 7.2 percentage points over the best SFT configuration on VideoDR and VDR-EE, respectively (Table~\ref{tab:training_tool_ablation}). Detailed tool-use analyses are outlined in Appendix~\ref{app:more_tool_use_analysis}.

\vspace{-2pt}
\subsection{Ablation}
Table~\ref{tab:training_tool_ablation} evaluates the contributions of SFT data composition and RL reward components on both benchmarks.

\begin{table}[H]
    \centering
    \caption{Ablation of SFT data and RL rewards. Accuracy is reported in \%; $\Delta$ is the gain in percentage points over the base model for SFT and over Reflection + Adaptive Data for RL.}
    \label{tab:training_tool_ablation}
    \small
    \renewcommand{\arraystretch}{1.15}
    \setlength{\tabcolsep}{7pt}
    \resizebox{0.75\linewidth}{!}{%
        \begin{tabular}{@{}lrrrr@{}}
            \toprule
            & \multicolumn{2}{c}{\textbf{VideoDR}} & \multicolumn{2}{c}{\textbf{VDR-EE}} \\
            \cmidrule(lr){2-3}\cmidrule(l){4-5}
            \textbf{Configuration} & Acc. & $\Delta$ & Acc. & $\Delta$ \\
            \midrule
            Base Model & 37.00 & -- & 40.00 & -- \\
            \midrule
            \multicolumn{5}{@{}l}{\textit{SFT data composition \quad ($\Delta$ vs. Base Model)}} \\
            \addlinespace[3pt]
            Base Data & 50.00 & +13.00 & 47.20 & +7.20 \\
            Adaptive Data & 48.00 & +11.00 & 46.00 & +6.00 \\
            Reflection Data & 51.00 & +14.00 & 47.60 & +7.60 \\
            Reflection + Adaptive Data & 52.00 & +15.00 & 48.00 & +8.00 \\
            \midrule
            \multicolumn{5}{@{}l}{\textit{RL reward components \quad ($\Delta$ vs. full SFT)}} \\
            \addlinespace[3pt]
            Accuracy + Format & 53.00 & +1.00 & 48.80 & +0.80 \\
            + Segment-Guided Evidence Reward & 57.00 & +5.00 & 50.80 & +2.80 \\
            + Tool-Type Normalization & 59.00 & +7.00 & 52.00 & +4.00 \\
            + Model-Conditioned Necessity Weight & \textbf{66.00} & \textbf{+14.00} & \textbf{55.20} & \textbf{+7.20} \\
            \bottomrule
        \end{tabular}}
\end{table}

\noindent\textbf{SFT data composition.} Training on base tool-use trajectories improves accuracy over the untrained model, establishing the benefit of learning to acquire evidence through tools. Adaptive data alone slightly lowers accuracy relative to base data, suggesting that learning to skip interactions does not by itself ensure that useful evidence is preserved. Reflection data improves performance by teaching the agent to revisit unreliable intermediate results. Combining reflection with adaptive data yields the strongest SFT configuration, reaching 52.00\% on VideoDR and 48.00\% on VDR-EE, or gains of 15.00 and 8.00 percentage points over the base model. These results support combining selective tool use with recovery from tool failures.

\noindent\textbf{RL reward components.} Starting from the combined SFT model, accuracy and format rewards produce only a small improvement, while segment-guided evidence rewards provide more direct feedback for useful evidence acquisition and raise accuracy to 57.00\% and 50.80\%. Tool-type normalization further improves these results to 59.00\% and 52.00\% by reducing the imbalance between tool types with different positive-reward rates. Finally, model-conditioned necessity weighting distinguishes useful evidence from tool calls that the model could bypass, increasing accuracy to 66.00\% and 55.20\%. This corresponds to total gains of 14.00 and 7.20 percentage points over SFT, while average interaction turns decrease from 13.67 to 11.67 on VideoDR and from 15.43 to 13.26 on VDR-EE relative to the variant without necessity weighting. Further ablations are reported in Appendix~\ref{app:more_ablations}.

\subsection{Training Curves}
\label{sec:training_reward_curves}

\begin{figure}[htbp]
    \centering
    \includegraphics[width=\linewidth]{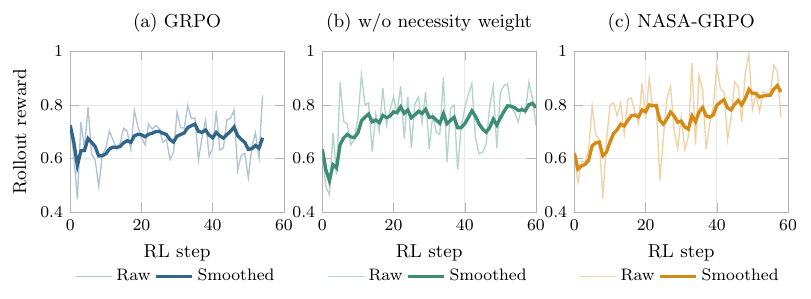}
    \caption{Training rollout rewards for (a) GRPO, (b) the variant without the necessity weight, and (c) NASA-GRPO. Pale and dark lines show raw and smoothed rewards. The curves in (b) are approximately digitized from the training-dashboard screenshot.}
    \label{fig:nasa_reward_curve}
\end{figure}

Figure~\ref{fig:nasa_reward_curve} compares the training rollout rewards of GRPO, the variant without the necessity weight, and NASA-GRPO. All three runs exhibit substantial fluctuations in raw reward, while the smoothed curves reveal different training trends. GRPO improves after an initial decline but shows no sustained upward trend in the later steps. Adding segment-guided evidence rewards and tool-type normalization without the necessity weight produces a clearer early improvement, followed by fluctuations around a relatively stable reward level. NASA-GRPO also improves rapidly at the beginning and maintains a higher smoothed reward toward the end of the displayed training window.

These trends are consistent with the ablation results in Table~\ref{tab:training_tool_ablation}: segment-guided credit assignment strengthens the learning signal beyond rollout-level feedback, while necessity weighting further improves the resulting policy. Evidence contribution alone can reward a successful tool call even when the model could recover the same information without it. The necessity weight reduces credit for such bypassable actions, directing the process signal toward evidence acquisition that the model actually needs. The reward curves provide a training diagnostic; the held-out accuracy gains and tool-necessity analysis in Appendix~\ref{app:tool_necessity_analysis} provide complementary evidence for improved tool selection.

\subsection{Qualitative Analysis}
\label{sec:qualitative_analysis}

\begin{figure}[H]
    \centering
    \includegraphics[width=\linewidth]{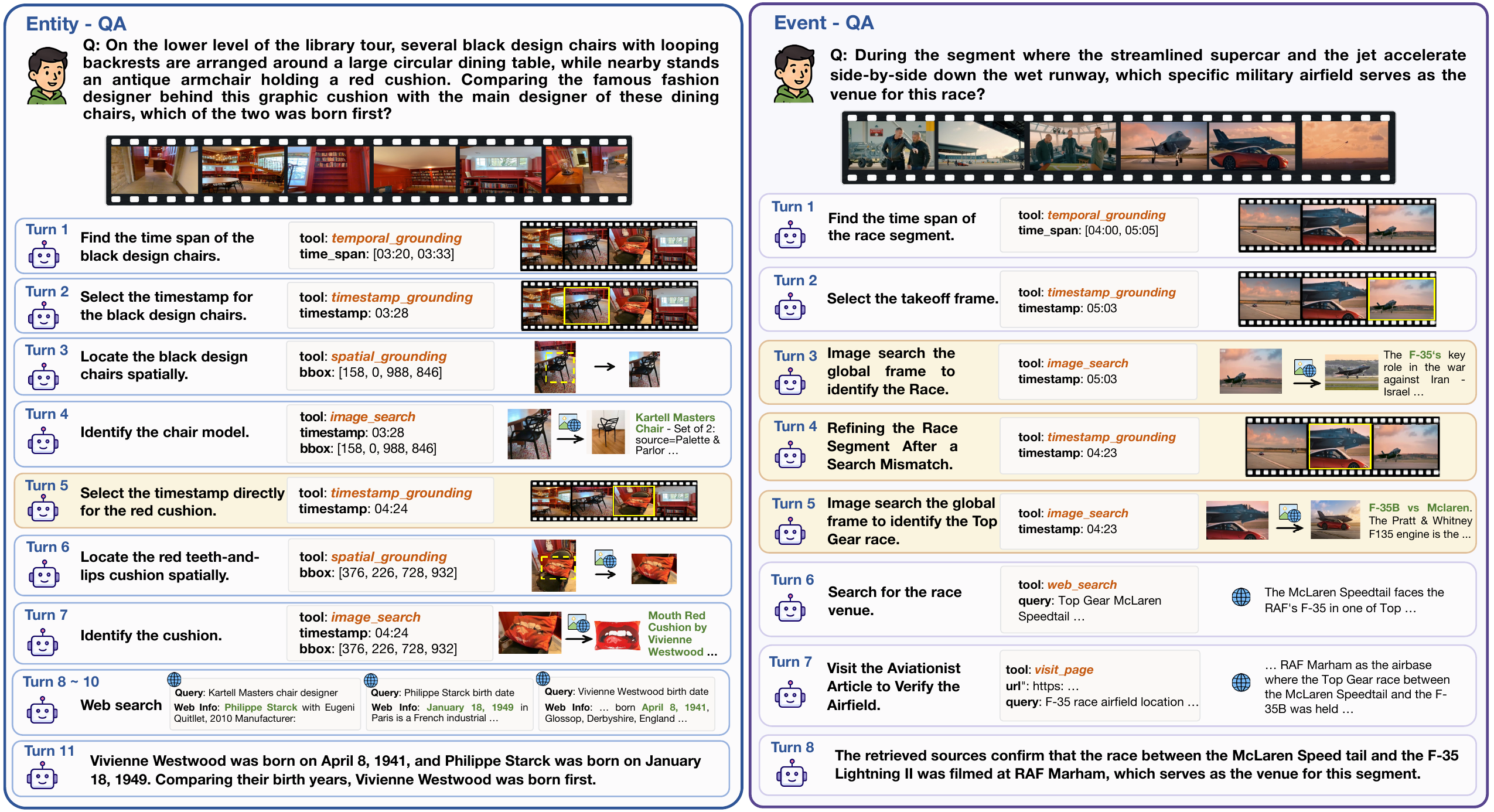}
    \vspace{-10pt}
    \caption{Qualitative examples of AdaVDR on entity-oriented and event-oriented questions.}
    \label{fig:qualitative_cases}
    \vspace{-10pt}
\end{figure}

Fig.~\ref{fig:qualitative_cases} illustrates how AdaVDR adapts its reasoning trajectory to the question and the evidence acquired during interaction. In the entity-oriented example, the model uses temporal grounding to locate the chairs. For the red cushion, the relevant timestamp can be directly identified from the current video context, so the model skips temporal grounding and proceeds to spatial grounding. It then identifies both objects before retrieving information about their designers and comparing their birth dates. This trajectory shows that AdaVDR can omit temporal grounding when the relevant timestamp can be identified directly. In the event-oriented example, the initial image search based on a takeoff frame returns mismatched evidence. The model therefore selects a more informative timestamp, identifies the race, and verifies the airfield through web search and page inspection. Together, the two examples demonstrate adaptive tool invocation and reflection-based recovery from unreliable intermediate retrieval results.

\vspace{-4pt}
\section{Conclusion}
\vspace{-4pt}

In this work, we study Video Deep Research and propose AdaVDR, an adaptive video deep research agent designed to handle diverse tool-use requirements. The developed data construction pipeline generates adaptive tool-use trajectories according to task requirements and model capabilities. The constructed training dataset initializes adaptive tool use and reflection through SFT, while NASA-GRPO further improves the policy through necessity-aware credit assignment. We also construct VDR-EE, a manually verified benchmark covering entity- and event-centric questions. The resulting AdaVDR selects necessary tools and corrects unreliable intermediate results when needed. Experiments demonstrate improvements in both accuracy and efficiency, with accuracy gains of 29.00 and 15.20 percentage points over the base model on VideoDR and VDR-EE, respectively.




\bibliography{ref}
\bibliographystyle{plainnat}

\appendix
\input{app.tex}

\end{document}

%% file: math_commands.tex
\usepackage{amsmath,amsfonts,bm}

\def\eqref#1{equation~\ref{#1}}

\def\1{\bm{1}}

\DeclareMathAlphabet{\mathsfit}{\encodingdefault}{\sfdefault}{m}{sl}
\SetMathAlphabet{\mathsfit}{bold}{\encodingdefault}{\sfdefault}{bx}{n}



%% file: app.tex
\clearpage
\section*{Appendix Contents}
\begingroup
\small
\setlength{\parskip}{2pt}
\noindent\hyperref[app:training_details]{\ref*{app:training_details}\quad Training Details}\dotfill\pageref*{app:training_details}\par
\noindent\hspace*{1.5em}\hyperref[app:training_parameters]{\ref*{app:training_parameters}\quad Training Parameters}\dotfill\pageref*{app:training_parameters}\par
\noindent\hyperref[app:proof_theorem1]{\ref*{app:proof_theorem1}\quad Proof of Theorem 1}\dotfill\pageref*{app:proof_theorem1}\par
\noindent\hyperref[app:more_results]{\ref*{app:more_results}\quad More Experimental Results}\dotfill\pageref*{app:more_results}\par
\noindent\hspace*{1.5em}\hyperref[app:vdr_ee_results]{\ref*{app:vdr_ee_results}\quad Results on VDR-EE}\dotfill\pageref*{app:vdr_ee_results}\par
\noindent\hspace*{1.5em}\hyperref[app:videodr_results]{\ref*{app:videodr_results}\quad Results on VideoDR}\dotfill\pageref*{app:videodr_results}\par
\noindent\hspace*{1.5em}\hyperref[app:image_deep_research]{\ref*{app:image_deep_research}\quad Image Deep Research}\dotfill\pageref*{app:image_deep_research}\par
\noindent\hspace*{1.5em}\hyperref[app:long_video_understanding]{\ref*{app:long_video_understanding}\quad Long-Video Understanding}\dotfill\pageref*{app:long_video_understanding}\par
\noindent\hyperref[app:more_tool_use_analysis]{\ref*{app:more_tool_use_analysis}\quad More Tool-Use Analysis}\dotfill\pageref*{app:more_tool_use_analysis}\par
\noindent\hspace*{1.5em}\hyperref[app:tool_effectiveness]{\ref*{app:tool_effectiveness}\quad Tool Effectiveness}\dotfill\pageref*{app:tool_effectiveness}\par
\noindent\hspace*{1.5em}\hyperref[app:tool_necessity_analysis]{\ref*{app:tool_necessity_analysis}\quad Tool Necessity Analysis}\dotfill\pageref*{app:tool_necessity_analysis}\par
\noindent\hyperref[app:training_dynamics]{\ref*{app:training_dynamics}\quad Training Dynamics}\dotfill\pageref*{app:training_dynamics}\par
\noindent\hspace*{1.5em}\hyperref[app:process_credit_dynamics]{\ref*{app:process_credit_dynamics}\quad Process Credit During Training}\dotfill\pageref*{app:process_credit_dynamics}\par
\noindent\hspace*{1.5em}\hyperref[app:necessity_weight_dynamics]{\ref*{app:necessity_weight_dynamics}\quad Effect of Model-Conditioned Necessity Weight on Tool Use}\dotfill\pageref*{app:necessity_weight_dynamics}\par
\noindent\hyperref[app:more_ablations]{\ref*{app:more_ablations}\quad More Ablations}\dotfill\pageref*{app:more_ablations}\par
\noindent\hspace*{1.5em}\hyperref[app:necessity_weight_ablation]{\ref*{app:necessity_weight_ablation}\quad Necessity Weight Ablation}\dotfill\pageref*{app:necessity_weight_ablation}\par
\noindent\hspace*{1.5em}\hyperref[app:process_advantage_ablation]{\ref*{app:process_advantage_ablation}\quad Process-Advantage Coefficient Ablation}\dotfill\pageref*{app:process_advantage_ablation}\par
\noindent\hspace*{1.5em}\hyperref[app:tool_ablation]{\ref*{app:tool_ablation}\quad Tool Ablation}\dotfill\pageref*{app:tool_ablation}\par
\noindent\hyperref[app:human_credit_agreement]{\ref*{app:human_credit_agreement}\quad Human--Credit Agreement}\dotfill\pageref*{app:human_credit_agreement}\par
\noindent\hyperref[app:prompts]{\ref*{app:prompts}\quad Prompts}\dotfill\pageref*{app:prompts}\par
\noindent\hspace*{1.5em}\hyperref[app:judge_prompt]{\ref*{app:judge_prompt}\quad LLM Judge Prompt}\dotfill\pageref*{app:judge_prompt}\par
\noindent\hspace*{1.5em}\hyperref[app:visual_anchor_prompt]{\ref*{app:visual_anchor_prompt}\quad Visual Anchor Extraction Prompt}\dotfill\pageref*{app:visual_anchor_prompt}\par
\noindent\hspace*{1.5em}\hyperref[app:qa_generation_prompt]{\ref*{app:qa_generation_prompt}\quad QA Generation Prompt}\dotfill\pageref*{app:qa_generation_prompt}\par
\noindent\hspace*{1.5em}\hyperref[app:initial_trajectory_generation_prompt]{\ref*{app:initial_trajectory_generation_prompt}\quad Initial Trajectory Generation Prompt}\dotfill\pageref*{app:initial_trajectory_generation_prompt}\par
\noindent\hspace*{1.5em}\hyperref[app:trajectory_refinement_prompt]{\ref*{app:trajectory_refinement_prompt}\quad Trajectory Refinement Prompt}\dotfill\pageref*{app:trajectory_refinement_prompt}\par
\noindent\hspace*{1.5em}\hyperref[app:necessity_prompts]{\ref*{app:necessity_prompts}\quad Model-Conditioned Tool Necessity Filtering Prompts}\dotfill\pageref*{app:necessity_prompts}\par
\noindent\hspace*{1.5em}\hyperref[app:retry_prompts]{\ref*{app:retry_prompts}\quad Reflection and Retry Prompts}\dotfill\pageref*{app:retry_prompts}\par
\endgroup
\clearpage

\section{Training Details}
\label{app:training_details}

\subsection{Training Parameters}
\label{app:training_parameters}

We train the model in two stages: supervised fine-tuning followed by reinforcement learning.

For SFT, we use five epochs, a learning rate of $2\times10^{-6}$, an effective batch size of 16, a maximum sequence length of 131,072, a LoRA rank of 8, sequence parallelism of 8, and gradient accumulation of 16.

The resulting SFT checkpoint initializes the policy for GRPO. Table~\ref{tab:rl_training_parameters} summarizes the RL configuration.

\begin{table}[htbp]
    \centering
    \caption{RL training configuration for NASA-GRPO.}
    \label{tab:rl_training_parameters}
    \small
    \setlength{\tabcolsep}{10pt}
    \renewcommand{\arraystretch}{1.05}
    \begin{tabular}{ll}
        \toprule
        \textbf{Parameter}                            & \textbf{Value}                     \\
        \midrule
        \rowcolor{gray!8}\multicolumn{2}{l}{\textit{Compute and parallelism}}              \\
        Backend                                       & Megatron                           \\
        GPUs                                          & 2 nodes $\times$ 8 GPUs (16 total) \\
        Training tensor parallelism                   & 4                                  \\
        Rollout tensor parallelism                    & 1                                  \\
        Pipeline parallelism                          & 1                                  \\
        Context parallelism                           & 1                                  \\
        Data parallelism                              & 4                                  \\
        Sequence parallelism                          & Enabled                            \\
        Dynamic batching                              & Enabled                            \\
        \rowcolor{gray!8}\multicolumn{2}{l}{\textit{Sampling and batching}}                \\
        Prompts per iteration                         & 8                                  \\
        Rollouts per prompt                           & 8                                  \\
        Trajectories per iteration                    & 64                                 \\
        Global batch size                             & 64                                 \\
        Micro batch size                              & 1                                  \\
        Rollout temperature                           & 1.0                                \\
        Maximum tool-interaction turns                & 20                                 \\
        \rowcolor{gray!8}\multicolumn{2}{l}{\textit{Optimization}}                         \\
        Optimizer                                     & Adam                               \\
        Learning-rate schedule                        & Constant                           \\
        Learning rate                                 & $1\times10^{-6}$                   \\
        Weight decay                                  & 0.1                                \\
        KL reward coefficient (\texttt{kl\_coef})     & 0                                  \\
        KL loss coefficient (\texttt{kl\_loss\_coef}) & 0                                  \\
        \bottomrule
    \end{tabular}
\end{table}

\section{Proof of Theorem 1}
\label{app:proof_theorem1}

\begin{definition}
    We define the \emph{residual task-relevant information} $\Gamma_M$ to measure the task-relevant information in $X$ that is not captured by $H_M$, formulated as
    \begin{equation}
        \Gamma_M := I(Y;X\mid H_M).
    \end{equation}
\end{definition}
\vspace{-1em}

\begin{definition}
    We define the \emph{misleadingness} $\delta$ of erroneous tool observations,
    which measures how strongly erroneous tool outputs distort the model prediction,
    formulated as
    \begin{equation}
        \delta :=
        \mathbb{E}\left[
            \operatorname{TV}\left(
            P(Y\mid H_M,Z),
            P(Y\mid H_M,Z;M)
            \right)
            \,\middle|\,
            C=1
            \right].
    \end{equation}
\end{definition}

\begin{definition}
    We define the \emph{base approximation error} $\xi_M$, which measures the
    mismatch between the ideal prediction and the actual prediction of model $M$
    given $H_M$, formulated as
    \begin{equation}
        \xi_M :=
        \mathbb{E}_{H_M}\left[
            D_{\mathrm{KL}}\left(
            P(Y\mid H_M)
            \,\Vert\,
            P(Y\mid H_M;M)
            \right)
            \right].
    \end{equation}
\end{definition}

\begin{lemma}
    We denote the \emph{base risk} $R_{\mathrm{base}}$ and the \emph{tool-augmented risk} $R_{\mathrm{tool}}$ as
    \begin{equation}
        \small
        \begin{aligned}
            R_{\mathrm{base}}
            = \mathbb{E}\left[-\log P(Y\mid H_M;M)\right],
            R_{\mathrm{tool}}
            = \mathbb{E}\left[-\log P(Y\mid H_M,Z;M)\right].
        \end{aligned}
    \end{equation}
    Let
    \begin{equation}
        \xi_{\mathrm{tool}}
        =
        \mathbb{E}_{H_M,Z}
        \left[
            D_{\mathrm{KL}}
            \left(
            P(Y\mid H_M,Z)
            \,\Vert\,
            P(Y\mid H_M,Z;M)
            \right)
            \right]
    \end{equation}
    denote the approximation error of the model with tool augmentation. Then,
    \begin{equation}
        R_{\mathrm{tool}}-R_{\mathrm{base}}
        =
        -I(Y;Z\mid H_M)
        +
        \xi_{\mathrm{tool}}-\xi_M.
        \label{eq:risk_decomposition}
    \end{equation}
\end{lemma}

\begin{proof}
    By the cross-entropy decomposition, the base risk satisfies
    \begin{align}
        R_{\mathrm{base}}
         & =
        \mathbb{E}\left[-\log P(Y\mid H_M;M)\right] \\
         & =
        H(Y\mid H_M)
        +
        \mathbb{E}_{H_M}
        \left[
            D_{\mathrm{KL}}
            \left(
            P(Y\mid H_M)
            \,\Vert\,
            P(Y\mid H_M;M)
            \right)
            \right]                          \\
         & =
        H(Y\mid H_M)+\xi_M.
    \end{align}
    Similarly, the tool-augmented risk satisfies
    \begin{align}
        R_{\mathrm{tool}}
         & =
        \mathbb{E}\left[-\log P(Y\mid H_M,Z;M)\right] \\
         & =
        H(Y\mid H_M,Z)
        +
        \mathbb{E}_{H_M,Z}
        \left[
            D_{\mathrm{KL}}
            \left(
            P(Y\mid H_M,Z)
            \,\Vert\,
            P(Y\mid H_M,Z;M)
            \right)
            \right]                           \\
         & =
        H(Y\mid H_M,Z)+\xi_{\mathrm{tool}}.
    \end{align}
    Therefore,
    \begin{align}
        R_{\mathrm{tool}}-R_{\mathrm{base}}
         & =
        H(Y\mid H_M,Z)-H(Y\mid H_M)
        +\xi_{\mathrm{tool}}-\xi_M \\
         & =
        -I(Y;Z\mid H_M)
        +\xi_{\mathrm{tool}}-\xi_M,
    \end{align}
    where the last equality follows from the definition of conditional mutual information.
\end{proof}

\begin{theorem}
    \label{theorem:theorem1_app}
    We define the \emph{base risk} $R_{\mathrm{base}}$ and the \emph{tool-augmented risk} $R_{\mathrm{tool}}$ as
    \begin{equation}
        \small
        \begin{aligned}
            R_{\mathrm{base}}
            = \mathbb{E}\left[-\log P(Y\mid H_M;M)\right],
            R_{\mathrm{tool}}
            = \mathbb{E}\left[-\log P(Y\mid H_M,Z;M)\right].
        \end{aligned}
    \end{equation}
    The difference between the base risk $R_{\mathrm{base}}$ and the tool-augmented risk $R_{\mathrm{tool}}$ satisfies
    \begin{equation}
        \small
        R_{\mathrm{tool}} - R_{\mathrm{base}}
        \ge
        2\epsilon\delta^2
        -
        \Gamma_M
        -
        \xi_M.
        \label{eq:tool_harm_bound_app}
    \end{equation}
    $\Gamma_M$  measures the necessary evidence in $X$ that is not captured by $H_M$.
    $\delta$  measures how strongly erroneous tool outputs distort the model prediction.
    $\xi_M$  measures the
    mismatch between the ideal prediction and the actual prediction of model $M$
    given $H_M$.
\end{theorem}

\begin{proof}
    According to the lemma, we have
    \begin{equation}
        R_{\mathrm{tool}}-R_{\mathrm{base}}
        =
        -I(Y;Z\mid H_M)
        +
        \xi_{\mathrm{tool}}
        -
        \xi_M,
        \label{eq:proof_decomposition}
    \end{equation}
    where
    \begin{equation}
        \xi_{\mathrm{tool}}
        =
        \mathbb{E}_{H_M,Z}
        \left[
            D_{\mathrm{KL}}
            \left(
            P(Y\mid H_M,Z)
            \,\Vert\,
            P(Y\mid H_M,Z;M)
            \right)
            \right].
    \end{equation}

    Since the tool observation $Z$ is generated from the original input $X$, we assume that
    $Y \rightarrow X \rightarrow Z$ forms a Markov chain. As $H_M=h_M(X)$ is a deterministic
    representation of $X$, the conditional data processing inequality gives
    \begin{equation}
        I(Y;Z\mid H_M)
        \le
        I(Y;X\mid H_M)
        =
        \Gamma_M.
        \label{eq:proof_information_bound}
    \end{equation}
    Therefore,
    \begin{equation}
        -I(Y;Z\mid H_M)
        \ge
        -\Gamma_M.
        \label{eq:proof_information_bound_negative}
    \end{equation}

    We next derive a lower bound for $\xi_{\mathrm{tool}}$.
    Let $C\in\{0,1\}$ indicate whether the tool output is erroneous, with
    $\Pr(C=1)=\epsilon$. By conditioning on $C$, we obtain
    \begin{align}
        \xi_{\mathrm{tool}}
         & =
        (1-\epsilon)
        \mathbb{E}\left[
                      D_{\mathrm{KL}}
                      \left(
                      P(Y\mid H_M,Z)
                      \,\Vert\,
                      P(Y\mid H_M,Z;M)
                      \right)
                      \middle| C=0
                      \right] \nonumber \\
         & \quad+
        \epsilon
        \mathbb{E}\left[
                      D_{\mathrm{KL}}
                      \left(
                      P(Y\mid H_M,Z)
                      \,\Vert\,
                      P(Y\mid H_M,Z;M)
                      \right)
                      \middle| C=1
                      \right] \nonumber \\
         & \ge
        \epsilon
        \mathbb{E}\left[
                      D_{\mathrm{KL}}
                      \left(
                      P(Y\mid H_M,Z)
                      \,\Vert\,
                      P(Y\mid H_M,Z;M)
                      \right)
                      \middle| C=1
                      \right].
    \end{align}

    By Pinsker's inequality,
    \begin{equation}
        D_{\mathrm{KL}}(P\Vert Q)
        \ge
        2\,\operatorname{TV}(P,Q)^2,
    \end{equation}
    and hence
    \begin{align}
        \xi_{\mathrm{tool}}
         & \ge
        2\epsilon
        \mathbb{E}\left[
                      \operatorname{TV}
                      \left(
                      P(Y\mid H_M,Z),
                      P(Y\mid H_M,Z;M)
                      \right)^2
                      \middle| C=1
                      \right] \nonumber \\
         & \ge
        2\epsilon
        \left(
        \mathbb{E}\left[
                      \operatorname{TV}
                      \left(
                      P(Y\mid H_M,Z),
                      P(Y\mid H_M,Z;M)
                      \right)
                      \middle| C=1
                      \right]
        \right)^2 \nonumber             \\
         & =
        2\epsilon\delta^2,
        \label{eq:proof_misleading_bound}
    \end{align}
    where the second inequality follows from Jensen's inequality and the last equality follows from the definition of $\delta$.

    Substituting Eq.~\ref{eq:proof_information_bound_negative} and
    Eq.~\ref{eq:proof_misleading_bound} into Eq.~\ref{eq:proof_decomposition}, we obtain
    \begin{align}
        R_{\mathrm{tool}}-R_{\mathrm{base}}
         & \ge
        -\Gamma_M
        +
        2\epsilon\delta^2
        -
        \xi_M \nonumber \\
         & =
        2\epsilon\delta^2
        -
        \Gamma_M
        -
        \xi_M.
    \end{align}
    Therefore,
    \begin{equation}
        R_{\mathrm{tool}}-R_{\mathrm{base}}
        \ge
        2\epsilon\delta^2-\Gamma_M-\xi_M,
    \end{equation}
    which completes the proof.
\end{proof}

\section{More Experimental Results}
\label{app:more_results}

We provide additional breakdowns of VDR-EE and VideoDR results by domain, video duration, and question difficulty, followed by qualitative examples of adaptive tool use and reflection.

\subsection{Results on VDR-EE}
\label{app:vdr_ee_results}

\begin{table}[htbp]
    \centering
    \caption{Domain-wise accuracy (\%) on VDR-EE under the \textbf{Direct} and \textbf{Agentic} settings.}
    \label{tab:our_bench_eval2}
    \renewcommand{\arraystretch}{1.30}
    \resizebox{1\textwidth}{!}{%
        \begin{tabular}{lcccccccc}
            \toprule
            \textbf{Model}                             & \multicolumn{7}{c}{\textbf{Domains}} & \textbf{Avg.}                                                                                                                                                                                                                                                                                                                \\
            \cmidrule(lr){2-8}
                                                       & \textbf{Culture}                     & \shortstack{\textbf{Entertain}\\\textbf{ment}} & \shortstack{\textbf{Indus}\\\textbf{trial}} & \textbf{News}                     & \multicolumn{1}{c}{\shortstack[c]{\textbf{Scene}\\\textbf{understanding}}} & \textbf{Sports}                    & \textbf{Science}                   &                                    \\
            \midrule
            \multicolumn{9}{c}{\textit{Direct}}                                                                                                                                                                                                                                                                                                                                                                              \\
            \midrule
            \rowcolor{gray!10}\multicolumn{9}{l}{\textit{Proprietary}}                                                                                                                                                                                                                                                                                                                                                       \\
            Gemini-3.1-Pro~\citep{team2023gemini}      & 38.89                                & 54.29                                          & 27.78                                       & 37.14                             & 63.89                                                                      & 50.00                              & 55.56                              & 46.80                              \\
            Gemini-3-Flash~\citep{team2023gemini}      & 44.44                                & 40.00                                          & 16.67                                       & 22.86                             & 66.67                                                                      & 44.44                              & 47.22                              & 40.40                              \\
            GPT-5.4~\citep{singh2025openai}            & 41.67                                & 14.29                                          & 16.67                                       & 22.86                             & 50.00                                                                      & 36.11                              & 41.67                              & 32.00                              \\
            \rowcolor{gray!10}\multicolumn{9}{l}{\textit{Open-source}}                                                                                                                                                                                                                                                                                                                                                        \\
            Qwen3-VL-8B-Instruct~\citep{bai2025qwen3}  & 2.78                                 & 14.29                                          & 11.11                                       & 8.57                              & 25.00                                                                      & 8.33                               & 16.67                              & 12.40                              \\
            Qwen3-VL-32B-Instruct~\citep{bai2025qwen3} & 11.11                                & 14.29                                          & 13.89                                       & 5.71                              & 36.11                                                                      & 11.11                              & 25.00                              & 16.80                              \\
            Qwen3.5-9B~\citep{team2026qwen3}           & 8.33                                 & 5.71                                           & 16.67                                       & 8.57                              & 25.00                                                                      & 8.33                               & 16.67                              & 12.80                              \\
            Qwen3.5-35B-A3B~\citep{team2026qwen3}      & 13.89                                & 5.71                                           & 5.56                                        & 11.43                             & 38.89                                                                      & 8.33                               & 27.78                              & 16.00                              \\
            \midrule
            \multicolumn{9}{c}{\textit{Agentic}}                                                                                                                                                                                                                                                                                                                                                                             \\
            \midrule
            \rowcolor{gray!10}\multicolumn{9}{l}{\textit{Proprietary}}                                                                                                                                                                                                                                                                                                                                                       \\
            Gemini-3.1-Pro~\citep{team2023gemini}      & 58.33                                & 74.29                                          & 50.00                                       & 51.43                             & 75.00                                                                      & 66.67                              & 75.00                              & 64.40                              \\
            Gemini-3-Flash~\citep{team2023gemini}      & 72.22                                & 51.43                                          & 41.67                                       & 48.57                             & 66.67                                                                      & 69.44                              & 66.67                              & 59.60                              \\
            GPT-5.4~\citep{singh2025openai}            & 47.22                                & 51.43                                          & 38.89                                       & 62.86                             & 66.67                                                                      & 44.44                              & 58.33                              & 52.80                              \\
            \rowcolor{gray!10}\multicolumn{9}{l}{\textit{Open-source}}                                                                                                                                                                                                                                                                                                                                                        \\
            Qwen3-VL-8B-Instruct~\citep{bai2025qwen3}  & 33.33                                & 28.57                                          & 25.00                                       & 28.57                             & 38.89                                                                      & 8.33                               & 36.11                              & 28.40                              \\
            Qwen3-VL-32B-Instruct~\citep{bai2025qwen3} & 50.00                                & 37.14                                          & 27.78                                       & 25.71                             & 52.78                                                                      & 11.11                              & 61.11                              & 38.00                              \\
            Qwen3.5-9B~\citep{team2026qwen3}           & 50.00                                & 34.29                                          & 30.56                                       & 40.00                             & 50.00                                                                      & 30.56                              & 44.44                              & 40.00                              \\
            Qwen3.5-35B-A3B~\citep{team2026qwen3}      & 55.56                                & 34.29                                          & 36.11                                       & 40.00                             & 55.56                                                                      & 36.11                              & 66.67                              & 46.40                              \\
            \rowcolor{green!10}\textbf{AdaVDR-9B (Qwen3.5-9B)}
                                                       & \textbf{63.89}                       & \textbf{57.14}                                 & \textbf{36.11}                              & \textbf{42.86}                    & \textbf{69.44}                                                             & \textbf{52.78}                     & \textbf{63.89}                     & \textbf{55.20}                     \\
            \textcolor{green!50!black}{$\Delta$ vs. Qwen3.5-9B}
                                                       & \textcolor{green!50!black}{+13.89}   & \textcolor{green!50!black}{+22.85}             & \textcolor{green!50!black}{+5.55}           & \textcolor{green!50!black}{+2.86} & \textcolor{green!50!black}{+19.44}                                         & \textcolor{green!50!black}{+22.22} & \textcolor{green!50!black}{+19.45} & \textcolor{green!50!black}{+15.20} \\
            \bottomrule
        \end{tabular}%
    }
    \renewcommand{\arraystretch}{1.0}
\end{table}

Table~\ref{tab:our_bench_eval2} reports domain-wise results on VDR-EE. AdaVDR-9B improves the overall accuracy of its Qwen3.5-9B base model from 40.00\% to 55.20\% and achieves gains across all seven domains. The largest improvements are observed in Entertainment and Sports, with gains of 22.85 and 22.22 percentage points, respectively.

\begin{table}[htbp]
    \centering
    \caption{Accuracy (\%) across full-video duration groups on VDR-EE under the \textbf{Direct} and \textbf{Agentic} settings.}
    \label{tab:our_bench_duration}
    \resizebox{0.7\textwidth}{!}{%
        \begin{tabular}{lrrrr}
            \toprule
            \textbf{Model}                             & \multicolumn{3}{c}{\textbf{Video Duration}} & \textbf{Avg.}                                                                                                               \\
            \cmidrule(lr){2-4}
                                                       & \shortstack{\textbf{Short}\\$\leq 5$ min}   & \shortstack{\textbf{Medium}\\$5<t<10$ min} & \shortstack{\textbf{Long}\\$\geq 10$ min} &                                    \\
            \midrule
            \multicolumn{5}{c}{\textit{Direct}}                                                                                                                                                                                    \\
            \midrule
            \rowcolor{gray!10}\multicolumn{5}{l}{\textit{Proprietary}}                                                                                                                                                             \\
            Gemini-3.1-Pro~\citep{team2023gemini}      & 50.65                                       & 52.50                                      & 38.71                                     & 46.80                              \\
            Gemini-3-Flash~\citep{team2023gemini}      & 41.56                                       & 48.75                                      & 32.26                                     & 40.40                              \\
            GPT-5.4~\citep{singh2025openai}            & 31.17                                       & 36.25                                      & 29.03                                     & 32.00                              \\
            \rowcolor{gray!10}\multicolumn{5}{l}{\textit{Open-source}}                                                                                                                                                              \\
            Qwen3-VL-8B-Instruct~\citep{bai2025qwen3}  & 6.49                                        & 15.00                                      & 15.05                                     & 12.40                              \\
            Qwen3-VL-32B-Instruct~\citep{bai2025qwen3} & 10.39                                       & 22.50                                      & 17.20                                     & 16.80                              \\
            Qwen3.5-9B~\citep{team2026qwen3}           & 11.69                                       & 18.75                                      & 8.60                                      & 12.80                              \\
            Qwen3.5-35B-A3B~\citep{team2026qwen3}      & 11.69                                       & 22.50                                      & 13.98                                     & 16.00                              \\
            \midrule
            \multicolumn{5}{c}{\textit{Agentic}}                                                                                                                                                                                   \\
            \midrule
            \rowcolor{gray!10}\multicolumn{5}{l}{\textit{Proprietary}}                                                                                                                                                             \\
            Gemini-3.1-Pro~\citep{team2023gemini}      & 68.83                                       & 62.50                                      & 62.37                                     & 64.40                              \\
            Gemini-3-Flash~\citep{team2023gemini}      & 71.43                                       & 63.75                                      & 46.24                                     & 59.60                              \\
            GPT-5.4~\citep{singh2025openai}            & 59.74                                       & 50.00                                      & 49.46                                     & 52.80                              \\
            \rowcolor{gray!10}\multicolumn{5}{l}{\textit{Open-source}}                                                                                                                                                              \\
            Qwen3-VL-8B-Instruct~\citep{bai2025qwen3}  & 31.17                                       & 32.50                                      & 22.58                                     & 28.40                              \\
            Qwen3-VL-32B-Instruct~\citep{bai2025qwen3} & 35.06                                       & 41.25                                      & 37.63                                     & 38.00                              \\
            Qwen3.5-9B~\citep{team2026qwen3}           & 32.47                                       & 51.25                                      & 36.56                                     & 40.00                              \\
            Qwen3.5-35B-A3B~\citep{team2026qwen3}      & 50.65                                       & 46.25                                      & 43.01                                     & 46.40                              \\
            \rowcolor{green!10}\textbf{AdaVDR-9B (Qwen3.5-9B)}
                                                       & \textbf{66.23}                              & \textbf{53.75}                             & \textbf{47.31}                            & \textbf{55.20}                     \\
            \textcolor{green!50!black}{$\Delta$ vs. Qwen3.5-9B}
                                                       & \textcolor{green!50!black}{+33.76}          & \textcolor{green!50!black}{+2.50}          & \textcolor{green!50!black}{+10.75}        & \textcolor{green!50!black}{+15.20} \\
            \bottomrule
        \end{tabular}%
    }
\end{table}

Table~\ref{tab:our_bench_duration} groups VDR-EE questions by full-video duration. AdaVDR-9B outperforms Qwen3.5-9B in all three groups, with gains of 33.76, 2.50, and 10.75 percentage points on short, medium, and long videos, respectively. The smaller gain on medium-length videos indicates that the benefit varies with video duration.

\subsection{Results on VideoDR}
\label{app:videodr_results}

\begin{table}[htbp]
    \centering
    \caption{Domain-wise accuracy (\%) on VideoDR under the \textbf{Workflow} and \textbf{Agentic} settings.}
    \label{tab:video_deepresearch_results}
    \resizebox{\textwidth}{!}{%
        \begin{tabular}{lrrrrrrr}
            \toprule
            \textbf{Model}                                      & \multicolumn{7}{c}{\textbf{VideoDR Domains}}                                                                                                                                                                                                                               \\
            \cmidrule(lr){2-8}
                                                                & \textbf{History}                             & \textbf{Geography}                 & \textbf{Culture}                   & \textbf{Economy}                   & \textbf{Technology}                & \textbf{Daily Life}                & \textbf{Avg.}                      \\
            \midrule
            \multicolumn{8}{c}{\textit{Workflow}}                                                                                                                                                                                                                                                                                            \\
            \midrule
            \rowcolor{gray!10}\multicolumn{8}{l}{\textit{Proprietary}}                                                                                                                                                                                                                                                                       \\
            Gemini-3-Pro-Preview~\citep{team2023gemini}         & 72.73                                        & 70.00                              & 80.00                              & 62.50                              & 64.29                              & 69.70                              & 69.00                              \\
            GPT-4o~\citep{hurst2024gpt}                         & 63.64                                        & 40.00                              & 33.33                              & 43.75                              & 42.86                              & 39.39                              & 42.00                              \\
            GPT-5.2~\citep{singh2025openai}                     & 72.73                                        & 70.00                              & 80.00                              & 56.25                              & 64.29                              & 72.73                              & 69.00                              \\
            \rowcolor{gray!10}\multicolumn{8}{l}{\textit{Open-source}}                                                                                                                                                                                                                                                                        \\
            Qwen3-Omni-30B-A3B-Instruct~\citep{xu2025qwen3}     & 36.36                                        & 30.00                              & 26.67                              & 43.75                              & 50.00                              & 36.36                              & 37.00                              \\
            InternVL3.5-14B~\citep{wang2025internvl3}           & 9.09                                         & 50.00                              & 20.00                              & 25.00                              & 21.43                              & 30.30                              & 27.00                              \\
            MiniCPM-V 4.5~\citep{yu2026minicpm}                 & 27.27                                        & 10.00                              & 46.67                              & 25.00                              & 14.29                              & 24.24                              & 25.00                              \\
            Qwen3-VL-8B-Instruct~\citep{bai2025qwen3}           & 36.36                                        & 30.00                              & 33.33                              & 31.25                              & 33.33                              & 18.18                              & 28.00                              \\
            Qwen3-VL-32B-Instruct~\citep{bai2025qwen3}          & 45.45                                        & 40.00                              & 46.67                              & 50.00                              & 26.67                              & 24.24                              & 36.00                              \\
            Qwen3.5-9B~\citep{team2026qwen3}                    & 54.55                                        & 20.00                              & 26.67                              & 50.00                              & 33.33                              & 30.30                              & 35.00                              \\
            Qwen3.5-35B-A3B~\citep{team2026qwen3}               & 63.64                                        & 20.00                              & 26.67                              & 56.25                              & 40.00                              & 33.33                              & 39.00                              \\
            \midrule
            \multicolumn{8}{c}{\textit{Agentic}}                                                                                                                                                                                                                                                                                             \\
            \midrule
            \rowcolor{gray!10}\multicolumn{8}{l}{\textit{Proprietary}}                                                                                                                                                                                                                                                                       \\
            Gemini-3-Pro-Preview~\citep{team2023gemini}         & 81.82                                        & 50.00                              & 86.67                              & 68.75                              & 85.71                              & 78.79                              & 76.00                              \\
            GPT-4o~\citep{hurst2024gpt}                         & 63.64                                        & 20.00                              & 53.33                              & 50.00                              & 35.71                              & 39.39                              & 43.00                              \\
            GPT-5.2~\citep{singh2025openai}                     & 90.91                                        & 70.00                              & 73.33                              & 56.25                              & 71.43                              & 66.67                              & 69.00                              \\
            \rowcolor{gray!10}\multicolumn{8}{l}{\textit{Open-source}}                                                                                                                                                                                                                                                                        \\
            Qwen3-Omni-30B-A3B-Instruct~\citep{xu2025qwen3}     & 54.55                                        & 40.00                              & 26.67                              & 43.75                              & 35.71                              & 33.33                              & 37.00                              \\
            InternVL3.5-14B~\citep{wang2025internvl3}           & 36.36                                        & 40.00                              & 26.67                              & 31.25                              & 28.57                              & 24.24                              & 30.00                              \\
            MiniCPM-V 4.5~\citep{yu2026minicpm}                 & 9.09                                         & 10.00                              & 26.67                              & 12.50                              & 14.29                              & 18.18                              & 16.00                              \\
            Qwen3-VL-8B-Instruct~\citep{bai2025qwen3}           & 36.36                                        & 40.00                              & 33.33                              & 31.25                              & 33.33                              & 21.21                              & 30.00                              \\
            Qwen3-VL-32B-Instruct~\citep{bai2025qwen3}          & 63.64                                        & 40.00                              & 46.67                              & 37.50                              & 40.00                              & 24.24                              & 38.00                              \\
            Qwen3.5-9B~\citep{team2026qwen3}                    & 54.55                                        & 20.00                              & 40.00                              & 43.75                              & 33.33                              & 33.33                              & 37.00                              \\
            Qwen3.5-35B-A3B~\citep{team2026qwen3}               & 63.64                                        & 20.00                              & 33.33                              & 62.50                              & 40.00                              & 36.36                              & 42.00                              \\
            Qwen3.5-397B-A17B~\citep{team2026qwen3}             & 54.55                                        & 70.00                              & 46.67                              & 81.25                              & 53.33                              & 48.48                              & 57.00                              \\
            \rowcolor{green!10}\textbf{AdaVDR-9B (Qwen3.5-9B)}  & \textbf{63.64}                               & \textbf{60.00}                     & \textbf{80.00}                     & \textbf{62.50}                     & \textbf{66.67}                     & \textbf{63.64}                     & \textbf{66.00}                     \\
            \textcolor{green!50!black}{$\Delta$ vs. Qwen3.5-9B} & \textcolor{green!50!black}{+9.09}            & \textcolor{green!50!black}{+40.00} & \textcolor{green!50!black}{+40.00} & \textcolor{green!50!black}{+18.75} & \textcolor{green!50!black}{+33.34} & \textcolor{green!50!black}{+30.31} & \textcolor{green!50!black}{+29.00} \\
            \bottomrule
        \end{tabular}%
    }
\end{table}

Table~\ref{tab:video_deepresearch_results} breaks down VideoDR accuracy by domain. AdaVDR-9B improves on the agentic Qwen3.5-9B base model in all six domains, raising overall accuracy from 37.00\% to 66.00\%. The largest gains occur in Geography and Culture (both +40.00 percentage points), while History shows the smallest gain (+9.09 points). Thus, the overall improvement is not driven by a single domain, although its magnitude varies across domains.

\begin{table}[htbp]
    \centering
    \caption{Accuracy (\%) across VideoDR difficulty levels under the \textbf{Agentic} setting.}
    \label{tab:difficulty_results}
    \resizebox{0.72\textwidth}{!}{%
        \begin{tabular}{lrrrr}
            \toprule
            \textbf{Model}                                      & \multicolumn{4}{c}{\textbf{Agentic}}                                                                                                                \\
            \cmidrule(lr){2-5}
                                                                & \textbf{Low}                         & \textbf{Medium}                    & \textbf{High}                      & \textbf{Avg.}                      \\
            \midrule
            \textbf{\#Samples}                                  & 32                                   & 36                                 & 32                                 & 100                                \\
            \midrule
            \rowcolor{gray!10}\multicolumn{5}{l}{\textit{Proprietary}}                                                                                                                                                \\
            Gemini-3-Pro-Preview~\citep{team2023gemini}         & 93.75                                & 69.44                              & 65.62                              & 76.00                              \\
            GPT-4o~\citep{hurst2024gpt}                         & 62.50                                & 38.89                              & 28.12                              & 43.00                              \\
            GPT-5.2~\citep{singh2025openai}                     & 84.38                                & 58.33                              & 65.62                              & 69.00                              \\
            \rowcolor{gray!10}\multicolumn{5}{l}{\textit{Open-source}}                                                                                                                                                 \\
            Qwen3-Omni-30B-A3B-Instruct~\citep{xu2025qwen3}     & 65.62                                & 27.78                              & 18.75                              & 37.00                              \\
            InternVL3.5-14B~\citep{wang2025internvl3}           & 46.88                                & 22.22                              & 21.88                              & 30.00                              \\
            MiniCPM-V 4.5~\citep{yu2026minicpm}                 & 18.75                                & 19.44                              & 9.38                               & 16.00                              \\
            Qwen3-VL-8B-Instruct~\citep{bai2025qwen3}           & 46.88                                & 27.78                              & 15.62                              & 30.00                              \\
            Qwen3-VL-32B-Instruct~\citep{bai2025qwen3}          & 46.88                                & 38.89                              & 28.12                              & 38.00                              \\
            Qwen3.5-9B~\citep{team2026qwen3}                    & 53.13                                & 33.33                              & 25.00                              & 37.00                              \\
            Qwen3.5-35B-A3B~\citep{team2026qwen3}               & 68.75                                & 41.67                              & 15.62                              & 42.00                              \\
            Qwen3.5-397B-A17B~\citep{team2026qwen3}             & 90.62                                & 58.33                              & 21.88                              & 57.00                              \\
            \rowcolor{green!10}\textbf{AdaVDR-9B (Qwen3.5-9B)}  & \textbf{84.38}                       & \textbf{61.11}                     & \textbf{53.13}                     & \textbf{66.00}                     \\
            \textcolor{green!50!black}{$\Delta$ vs. Qwen3.5-9B} & \textcolor{green!50!black}{+31.25}   & \textcolor{green!50!black}{+27.78} & \textcolor{green!50!black}{+28.13} & \textcolor{green!50!black}{+29.00} \\
            \bottomrule
        \end{tabular}%
    }
\end{table}

Table~\ref{tab:difficulty_results} reports VideoDR results across difficulty levels. AdaVDR-9B reaches 66.00\% overall accuracy, outperforming its Qwen3.5-9B base model by 29.00 percentage points. The improvements are consistent across the Low-, Medium-, and High-difficulty subsets, with gains of 31.25, 27.78, and 28.13 points, respectively.

We also examine whether AdaVDR generalizes beyond video deep research in two related settings.

\subsection{Image Deep Research}
\label{app:image_deep_research}

We evaluate AdaVDR on image-based research tasks with video-specific grounding tools disabled. Table~\ref{tab:image_deep_research} reports results on five image deep-research benchmarks.

\begin{table}[htbp]
    \centering
    \caption{Accuracy (\%) on five image deep-research benchmarks.}
    \label{tab:image_deep_research}
    \footnotesize
    \setlength{\tabcolsep}{4.5pt}
    \renewcommand{\arraystretch}{1.03}
    \begin{tabular*}{\textwidth}{@{\extracolsep{\fill}}lrrrrrr}
        \toprule
        \textbf{Model} & \textbf{SimpleVQA} & \textbf{MMSearch} & \textbf{LiveVQA} & \textbf{BC-VL} & \textbf{FVQA} & \textbf{Avg.} \\
        \midrule
        \rowcolor{gray!8}\multicolumn{7}{l}{\textit{Agentic workflow}} \\
        GPT-5 & 67.3 & 62.7 & -- & 57.6 & 62.0 & -- \\
        Gemini-2.5-Pro & 74.3 & 55.7 & -- & 42.3 & 65.0 & -- \\
        Gemini-3.1-Pro & -- & 86.1 & 76.6 & 64.1 & 84.0 & -- \\
        Claude-4-Sonnet & -- & 67.2 & -- & 48.6 & 69.0 & -- \\
        Claude-4.6-Opus & -- & 76.2 & 67.4 & 48.3 & 74.5 & -- \\
        Kimi-K2.5 & -- & 76.6 & 76.6 & 50.3 & 76.5 & -- \\
        \midrule
        \rowcolor{gray!8}\multicolumn{7}{l}{\textit{Multimodal deep-research models}} \\
        Visual-ARFT-7B & 42.4 & 34.5 & 25.4 & 16.5 & 41.7 & 32.1 \\
        MMSearch-R1-7B & 57.4 & 53.8 & 48.4 & 20.9 & 58.4 & 47.8 \\
        DeepEyes-V2-7B & 59.4 & 63.7 & -- & -- & 60.6 & -- \\
        WebWatcher-7B & 54.3 & 49.1 & 51.2 & 21.2 & -- & -- \\
        Qwen3-VL-8B & 52.0 & 37.4 & 50.6 & 27.9 & 58.7 & 45.3 \\
        SenseSearch-8B & 61.7 & 67.4 & 56.2 & 35.1 & 67.1 & 57.5 \\
        OpenSearch-VL-8B & 71.6 & 64.5 & 59.6 & 37.6 & 71.5 & 61.0 \\
        \addlinespace[2pt]
        Qwen3-VL-30B & 55.1 & 44.2 & 62.0 & 34.1 & 63.0 & 51.7 \\
        \addlinespace[2pt]
        WebWatcher-32B & 59.0 & 55.3 & 58.7 & 27.0 & -- & -- \\
        Qwen3.5-9B (Agentic) & 56.8 & 41.6 & 55.4 & 31.7 & 63.2 & 49.7 \\
        Qwen3.5-27B (Agentic) & 64.3 & 59.1 & 50.6 & 39.4 & 67.3 & 56.1 \\
        \midrule
        \rowcolor{gray!12}\textbf{AdaVDR-9B} & \textbf{72.0} & \textbf{70.7} & \textbf{64.5} & \textbf{39.5} & \textbf{74.0} & \textbf{64.1} \\
        \bottomrule
    \end{tabular*}
\end{table}

\subsection{Long-Video Understanding}
\label{app:long_video_understanding}

We evaluate AdaVDR-9B on three long-video understanding benchmarks, as shown in Table~\ref{tab:video_understanding_backbone}.

\begin{table}[htbp]
    \centering
    \caption{Video question-answering accuracy (\%).}
    \label{tab:video_understanding_backbone}
    \scriptsize
    \setlength{\tabcolsep}{3pt}
    \resizebox{0.7\textwidth}{!}{%
        \begin{tabular}{lrrrr}
            \toprule
            \textbf{Model}                       & \textbf{VideoMME} & \textbf{MV-Bench} & \textbf{LongVideoBench} & \textbf{Avg.} \\
            \midrule
            InternVL3.5-8B                       & 66.0              & 72.1              & 62.1                    & 66.7          \\
            Keye-VL-1.5-8B                       & 73.0              & 56.9              & 66.0                    & 65.3          \\
            Eagle2.5-8B                          & 72.4              & 74.8              & 66.4                    & 71.2          \\
            MiniCPM-V-4.5-8B                     & 67.9              & 60.5              & 63.9                    & 64.1          \\
            LLaVA-OneVision-2-8B                 & 71.9              & 66.2              & 66.9                    & 68.3          \\
            Qwen3.5-9B                           & 71.1              & 68.2              & 64.6                    & 68.0          \\
            \midrule
            Video-RTS                            & 63.0              & --                & 56.6                    & --            \\
            Video-MTR                            & 59.0              & --                & 56.4                    & --            \\
            LOVE-R1                              & 64.3$^*$          & --                & 58.0$^*$                & --            \\
            Conan                                & 60.5              & --                & 56.6                    & --            \\
            LongVT                               & 64.3              & --                & --                      & --            \\
            Video-Zoomer                         & 64.6$^*$          & --                & 55.9$^*$                & --            \\
            Video-o3                             & 66.5              & --                & 60.5                    & --            \\
            \midrule
            \rowcolor{gray!12}\textbf{AdaVDR-9B} & 72.8              & 74.9              & 68.2                    & 72.0          \\
            \bottomrule
        \end{tabular}}
\end{table}

\section{More Tool-Use Analysis}
\label{app:more_tool_use_analysis}

\subsection{Tool Effectiveness}
\label{app:tool_effectiveness}

We compare SFT and NASA-GRPO on the same evaluation questions using the visual and text segments defined in Sec.~\ref{section:training}. A visual segment is useful when its grounding, with image search if used, provides the target visual information; a timestamp-only segment can therefore be useful. A text segment is useful when its Web search and associated page visits, or a standalone page visit, recover the target fact. We then inspect whether the model treats the target as established despite an unsuccessful segment, either by answering from prior knowledge without retrieved support or by relying on an incorrect tool result.

We manually inspect trajectories for 100 VideoDR questions. For VDR-EE, a VLM judge applies the same two checks to each segment. For useful segments, we also bypass the segment and ask the evaluated model to recover its target from the preceding context; we count it as necessary if the model cannot do so. Table~\ref{tab:retrieval_effectiveness} reports percentages computed from pooled segment counts. Necessary segments are reported as a percentage of useful segments. False acceptance is the share of all segments that fail to obtain useful information but whose target the model nevertheless treats as established.

\begin{table}[htbp]
    \centering
    \caption{Tool effectiveness and necessity for visual and text segments on 100 manually reviewed VideoDR questions and VLM-judged VDR-EE questions. Useful segments and false acceptance use all segments as the denominator; necessary segments use useful segments as the denominator.}
    \label{tab:retrieval_effectiveness}
    \small
    \setlength{\tabcolsep}{5pt}
    \resizebox{0.85\textwidth}{!}{%
        \begin{tabular}{lllrrr}
            \toprule
            \textbf{Benchmark} & \textbf{Model} & \textbf{Segment} & \shortstack{\textbf{Useful segments}\\\textbf{(\%)}} & \shortstack{\textbf{Necessary among useful}\\\textbf{(\%)}} & \shortstack{\textbf{False acceptance}\\\textbf{(\%)}} \\
            \midrule
            VideoDR & SFT       & Visual & 58.4 & 68.9 & 27.6 \\ 
                    &           & Text   & 64.4 & 71.4 & 19.4 \\ 
                    & NASA-GRPO & Visual & 76.1 & 84.6 & 12.8 \\ 
                    &           & Text   & 77.8 & 85.2 & 11.7 \\ 
            VDR-EE  & SFT       & Visual & 60.7 & 69.7 & 23.4 \\ 
                    &           & Text   & 64.6 & 68.8 & 21.6 \\ 
                    & NASA-GRPO & Visual & 74.1 & 83.9 & 13.6 \\ 
                    &           & Text   & 78.4 & 85.0 & 9.4 \\ 
            \bottomrule
        \end{tabular}}
\end{table}

\begin{figure}[htbp]
    \centering
    \includegraphics[width=0.98\linewidth]{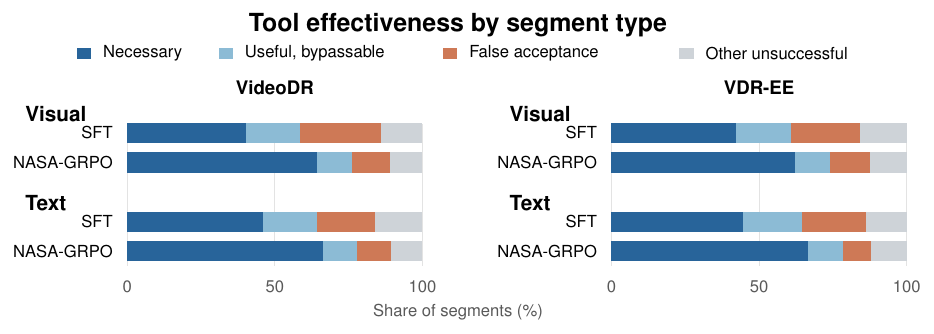}
    \caption{Tool effectiveness by segment type on VideoDR and VDR-EE. Necessary and useful but bypassable segments partition useful segments; false acceptance and other unsuccessful segments partition the remaining segments.}
    \label{fig:tool_effectiveness}
\end{figure}

Table~\ref{tab:retrieval_effectiveness} and Fig.~\ref{fig:tool_effectiveness}, together with Table~\ref{tab:tool_analysis}, support the view that NASA-GRPO improves evidence acquisition rather than merely increasing tool calls, because it has higher useful- and necessary-segment rates and lower false-acceptance rates than SFT for both visual and text segments. Its higher visual usefulness despite fewer grounding calls is also consistent with selecting inputs that better support image search.

\begin{figure}[htbp]
    \centering
    \includegraphics[width=\linewidth,height=0.76\textheight,keepaspectratio]{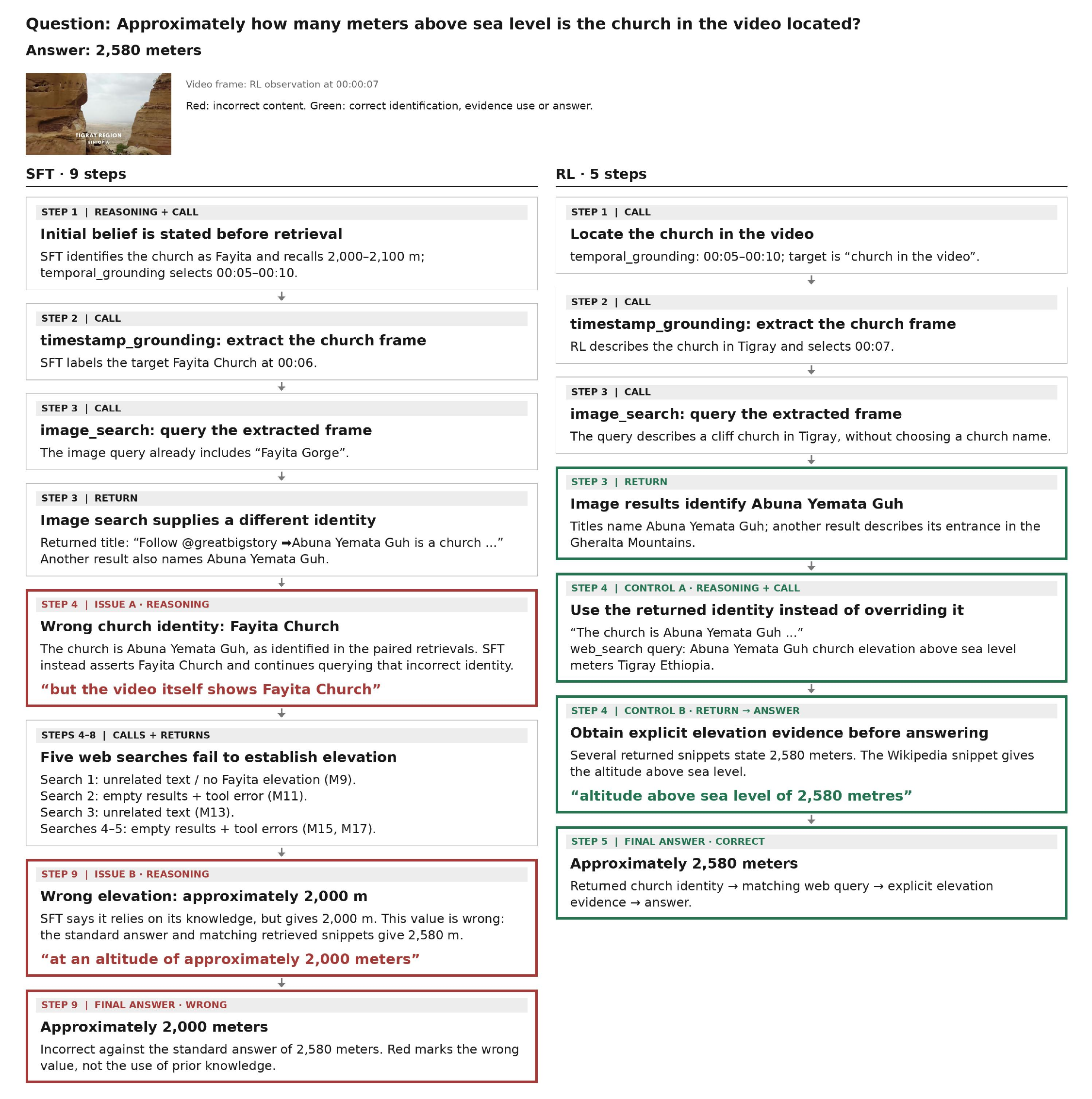}
    \caption{A VideoDR case illustrating unsupported answering after failed evidence acquisition. SFT retains an incorrect church identity despite the image-search result and answers approximately 2,000 m without supporting elevation evidence. The RL trajectory uses the retrieved church identity and elevation evidence to answer approximately 2,580 m.}
    \label{fig:sft_false_acceptance_case}
\end{figure}

Fig.~\ref{fig:sft_false_acceptance_case} illustrates a false-acceptance failure in one SFT trajectory. Although image search returns the correct church identity, SFT retains an earlier, incorrect identification and gives an unsupported elevation after subsequent searches fail to establish the target fact. The RL trajectory instead follows the retrieved identity and obtains explicit elevation evidence before answering.

\subsection{Tool Necessity Analysis}
\label{app:tool_necessity_analysis}

We compare the RL variants after tool-type normalization, with and without the model-conditioned necessity weight (Table~\ref{tab:training_tool_ablation}), on the same VideoDR and VDR-EE questions under identical decoding settings. We identify useful evidence-acquisition segments using the criteria in Sec.~\ref{app:tool_effectiveness} and apply the model-conditioned recovery criterion of Eq.~\ref{eq:necessity_weight}. A segment is necessary when it is useful and its target information cannot be recovered after bypassing it. We assess tool calls likewise: a call is necessary when it retrieves useful target information that the evaluated model cannot recover after bypassing the call; otherwise, it is redundant.

Table~\ref{tab:tool_necessity_analysis} reports necessary segments as a percentage of all segments and necessary and redundant calls as percentages of all calls, with counts pooled across questions. This analysis complements the answer-accuracy ablation in Table~\ref{tab:training_tool_ablation} by measuring whether the necessity weight shifts tool use toward evidence and calls that provide otherwise unavailable information.

\begin{table}[htbp]
    \centering
    \caption{Necessity of evidence-acquisition segments and tool calls on VideoDR and VDR-EE. Segment percentages use all segments; call percentages use all tool calls.}
    \label{tab:tool_necessity_analysis}
    \small
    \setlength{\tabcolsep}{5pt}
    \resizebox{0.75\textwidth}{!}{%
        \begin{tabular}{llrrr}
            \toprule
            \textbf{Benchmark} & \textbf{Necessity weight} & \shortstack{\textbf{Necessary}\\\textbf{segments (\%)}} & \shortstack{\textbf{Necessary}\\\textbf{calls (\%)}} & \shortstack{\textbf{Redundant}\\\textbf{calls (\%)}} \\
            \midrule
            VideoDR            & Without                   & 54.9 & 54.6 & 45.4 \\
                               & With                      & 65.9 & 76.2 & 23.8 \\
            VDR-EE             & Without                   & 50.8 & 53.2 & 46.8 \\
                               & With                      & 65.9 & 74.5 & 25.5 \\
            \bottomrule
        \end{tabular}}
\end{table}

\begin{figure}[htbp]
    \centering
    \includegraphics[width=0.98\linewidth]{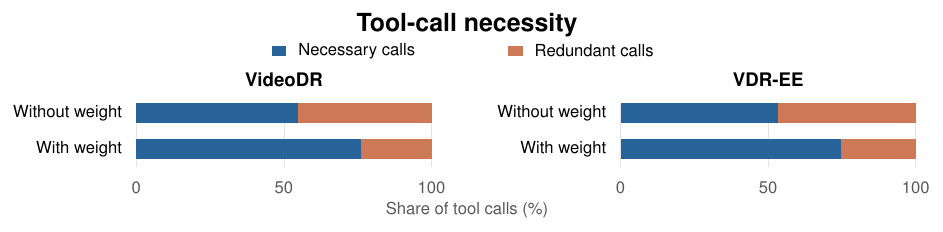}
    \caption{Necessary and redundant tool calls on VideoDR and VDR-EE, with and without the model-conditioned necessity weight. Shares are computed over all tool calls.}
    \label{fig:tool_necessity}
\end{figure}

Table~\ref{tab:tool_necessity_analysis} shows that necessity-aware training raises the share of necessary segments from 54.9\% to 65.9\% on VideoDR and from 50.8\% to 65.9\% on VDR-EE. The necessary tool-call share rises from 54.6\% to 76.2\% and from 53.2\% to 74.5\%, respectively, as illustrated in Fig.~\ref{fig:tool_necessity}. Despite the corresponding reduction in redundant calls, Table~\ref{tab:training_tool_ablation} shows that accuracy rises from 59.00\% to 66.00\% on VideoDR and from 52.00\% to 55.20\% on VDR-EE. This pattern indicates that the weight improves tool selection rather than simply reducing tool use.

\section{Training Dynamics}
\label{app:training_dynamics}

We examine the evolution of process credits and the effect of the model-conditioned necessity weight on tool use.

\subsection{Process Credit During Training}
\label{app:process_credit_dynamics}

\begin{table}[htbp]
    \centering
    \caption{Tool-level process credits during RL training with the model-conditioned necessity weight. Positive-credit calls receive 0.5 or 1.0; no-credit calls receive 0.}
    \label{tab:training_credit_dynamics}
    \small
    \setlength{\tabcolsep}{8pt}
    \begin{tabular}{lrrrrrr}
        \toprule
        \textbf{Rollouts} & \shortstack{\textbf{All tool}\\\textbf{calls}} & \shortstack{\textbf{Credit 1.0}\\\textbf{calls}} & \shortstack{\textbf{Credit 0.5}\\\textbf{calls}} & \shortstack{\textbf{Positive-credit}\\\textbf{calls}} & \shortstack{\textbf{Positive-credit}\\\textbf{share (\%)}} & \shortstack{\textbf{No-credit}\\\textbf{share (\%)}} \\
        \midrule
        0--9              & 3775                                           & 589                                              & 213                                              & 802                                                   & 21.2                                                       & 78.8                                                 \\
        10--19            & 3878                                           & 652                                              & 374                                              & 1026                                                  & 26.5                                                       & 73.5                                                 \\
        20--29            & 4947                                           & 878                                              & 362                                              & 1240                                                  & 25.1                                                       & 74.9                                                 \\
        30--39            & 5059                                           & 922                                              & 616                                              & 1538                                                  & 30.4                                                       & 69.6                                                 \\
        40--49            & 4753                                           & 923                                              & 582                                              & 1505                                                  & 31.7                                                       & 68.3                                                 \\
        \bottomrule
    \end{tabular}
\end{table}

\begin{figure}[H]
    \centering
    \includegraphics[width=0.76\linewidth]{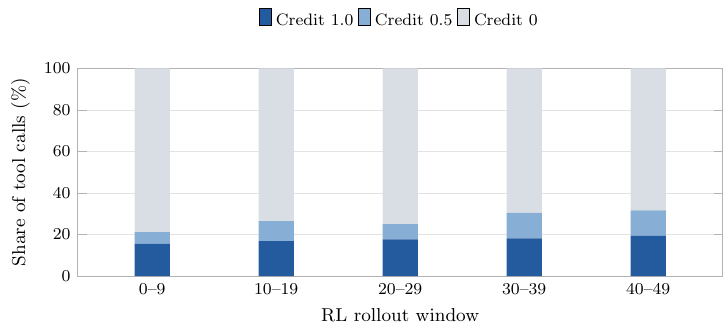}
    \caption{Process-credit composition across RL rollout windows. Each bar shows the share of calls assigned credit 0, 0.5, or 1.0, using the counts in Table~\ref{tab:training_credit_dynamics}.}
    \label{fig:process_credit_trend}
\end{figure}

Table~\ref{tab:training_credit_dynamics} and Fig.~\ref{fig:process_credit_trend} show that the share of calls receiving positive process credit rises overall from 21.2\% to 31.7\%, while the share receiving no credit falls from 78.8\% to 68.3\%, despite a small reversal in the middle of training. Both full- and half-credit calls increase in share, with full-credit calls remaining more numerous throughout. Together, these trends suggest that the model becomes more effective at acquiring useful external evidence as training progresses.

\subsection{Effect of Model-Conditioned Necessity Weight on Tool Use}
\label{app:necessity_weight_dynamics}

\begin{table}[htbp]
    \centering
    \caption{Tool-call distribution in a training snapshot with and without the model-conditioned necessity weight.}
    \label{tab:necessity_tool_dynamics_train}
    \small
    \setlength{\tabcolsep}{7pt}
    \begin{tabular}{lrrrrr}
        \toprule
        \textbf{Tool type}  & \multicolumn{2}{c}{\textbf{Without necessity weight}} & \multicolumn{2}{c}{\textbf{With necessity weight}} & \textbf{$\Delta$ Calls}                                      \\
        \cmidrule(lr){2-3}\cmidrule(lr){4-5}
                            & \textbf{Calls}                                        & \textbf{Share (\%)}                                & \textbf{Calls}          & \textbf{Share (\%)} &              \\
        \midrule
        temporal grounding  & 62                                                    & 10.42                                              & 65                      & 11.44               & +3           \\
        timestamp grounding & 60                                                    & 10.08                                              & 65                      & 11.44               & +5           \\
        spatial grounding   & 6                                                     & 1.01                                               & 4                       & 0.70                & -2           \\
        image search        & 54                                                    & 9.08                                               & 70                      & 12.32               & +16          \\
        web search          & 344                                                   & 57.82                                              & 200                     & 35.21               & -144         \\
        page visit          & 69                                                    & 11.60                                              & 164                     & 28.87               & +95          \\
        \midrule
        \textbf{Total}      & \textbf{595}                                          & \textbf{100}                                       & \textbf{568}            & \textbf{100}        & \textbf{-27} \\
        \bottomrule
    \end{tabular}
\end{table}

\begin{figure}[htbp]
    \centering
    \includegraphics[width=0.88\linewidth]{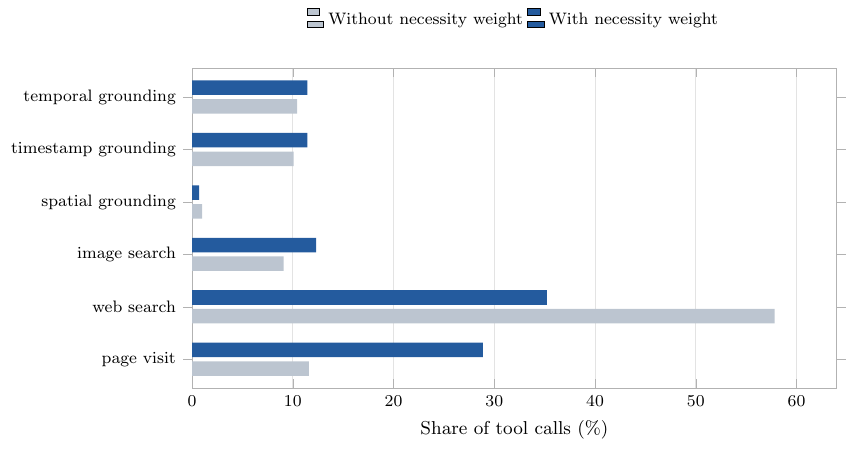}
    \caption{Tool-use distribution in a training snapshot with and without the model-conditioned necessity weight. Bars show each tool's share of all calls.}
    \label{fig:tool_mix_comparison_train}
\end{figure}

In this training snapshot, the model without the model-conditioned necessity weight makes more tool calls than the weighted model (595 vs. 568; Table~\ref{tab:necessity_tool_dynamics_train}). Fig.~\ref{fig:tool_mix_comparison_train} highlights the shift in tool use: web search falls from 57.82\% to 35.21\% of calls, while page visits rise from 11.60\% to 28.87\% and image search rises from 9.08\% to 12.32\%. This shift toward visual retrieval and page inspection complements the accuracy gains in Table~\ref{tab:training_tool_ablation}.

\section{More Ablations}
\label{app:more_ablations}

We examine the two reward coefficients separately and assess the contribution of each available tool.

\subsection{Necessity Weight Ablation}
\label{app:necessity_weight_ablation}

For bypassable actions, we vary the necessity weight while keeping the weight of necessary actions at 1. The default bypassable-action weight is 0.5 (Eq.~\ref{eq:necessity_weight}); all other training settings remain fixed.

\begin{table}[htbp]
    \centering
    \caption{Ablation of the necessity weight assigned to bypassable tool calls. The default is 0.5.}
    \label{tab:necessity_weight_ablation}
    \small
    \setlength{\tabcolsep}{9pt}
    \begin{tabular}{ccccc}
        \toprule
        \multirow{2}{*}{\shortstack{\textbf{Bypassable-call}\\\textbf{weight}}} & \multicolumn{2}{c}{\textbf{VideoDR}} & \multicolumn{2}{c}{\textbf{VDR-EE}}                                                       \\
        \cmidrule(lr){2-3}\cmidrule(lr){4-5}
                                                               & \textbf{Acc. (\%)}                   & \textbf{Avg. calls}                 & \textbf{Acc. (\%)}       & \textbf{Avg. calls}      \\
        \midrule
        0.2                                                    & \textcolor{black}{61.00}             & \textcolor{black}{9.10}             & \textcolor{black}{50.80} & \textcolor{black}{10.50} \\
        0.3                                                    & \textcolor{black}{63.00}             & \textcolor{black}{9.55}             & \textcolor{black}{52.40} & \textcolor{black}{11.05} \\
        0.4                                                    & \textcolor{black}{64.00}             & \textcolor{black}{10.10}            & \textcolor{black}{54.00} & \textcolor{black}{11.65} \\
        \textbf{0.5 (default)}                                 & \textcolor{black}{66.00}             & \textcolor{black}{10.67}            & \textcolor{black}{55.20} & \textcolor{black}{12.26} \\
        0.6                                                    & \textcolor{black}{62.00}             & \textcolor{black}{11.75}            & \textcolor{black}{52.00} & \textcolor{black}{13.40} \\
        \bottomrule
    \end{tabular}
\end{table}

The comparisons suggest a trade-off: smaller weights weaken the process signal even for bypassable calls that return correct evidence, whereas larger weights narrow the reward gap between necessary and bypassable calls and may encourage redundant tool use.

\subsection{Process-Advantage Coefficient Ablation}
\label{app:process_advantage_ablation}

We vary the coefficient $\lambda$ that scales the process advantage in Eq.~\ref{eq:final_advantage}, keeping the necessity weights fixed at their default values. The default is $\lambda=0.3$.

\begin{table}[htbp]
    \centering
    \caption{Ablation of the process-advantage coefficient $\lambda$. The default is 0.3.}
    \label{tab:process_advantage_ablation}
    \small
    \setlength{\tabcolsep}{9pt}
    \begin{tabular}{ccccc}
        \toprule
        \textbf{$\lambda$}     & \multicolumn{2}{c}{\textbf{VideoDR}} & \multicolumn{2}{c}{\textbf{VDR-EE}}                                                       \\
        \cmidrule(lr){2-3}\cmidrule(lr){4-5}
                               & \textbf{Acc. (\%)}                   & \textbf{Avg. calls}                 & \textbf{Acc. (\%)}       & \textbf{Avg. calls}      \\
        \midrule
        0.2                    & \textcolor{black}{62.00}             & \textcolor{black}{8.40}             & \textcolor{black}{51.60} & \textcolor{black}{10.00} \\
        \textbf{0.3 (default)} & \textcolor{black}{66.00}             & \textcolor{black}{10.67}            & \textcolor{black}{55.20} & \textcolor{black}{12.26} \\
        0.4                    & \textcolor{black}{64.00}             & \textcolor{black}{12.00}            & \textcolor{black}{53.60} & \textcolor{black}{13.60} \\
        0.5                    & \textcolor{black}{61.00}             & \textcolor{black}{13.40}            & \textcolor{black}{51.20} & \textcolor{black}{15.00} \\
        \bottomrule
    \end{tabular}
\end{table}

The values peak at $\lambda=0.3$. A smaller coefficient may underweight evidence-acquisition actions, reducing useful retrieval. A larger coefficient may overemphasize process rewards, leading to more tool calls without better answers.

\subsection{Tool Ablation}
\label{app:tool_ablation}

We keep the trained AdaVDR model fixed and disable one tool at a time during inference, leaving the other tools available. Table~\ref{tab:tool_ablation} compares each setting with the full tool set; $\Delta$ denotes the accuracy change in percentage points.

\begin{table}[htbp]
    \centering
    \caption{Inference-time tool ablation. Each row independently removes only the named tool from the full model; all other tools remain available. Full-model accuracies match Table~\ref{tab:our_bench_eval1}.}
    \label{tab:tool_ablation}
    \small
    \setlength{\tabcolsep}{9pt}
    \begin{tabular}{lcccc}
        \toprule
        \textbf{Configuration}       & \multicolumn{2}{c}{\textbf{VideoDR}} & \multicolumn{2}{c}{\textbf{VDR-EE}}                                                          \\
        \cmidrule(lr){2-3}\cmidrule(lr){4-5}
                                     & \textbf{Acc. (\%)}                   & \textbf{$\Delta$}                   & \textbf{Acc. (\%)}       & \textbf{$\Delta$}           \\
        \midrule
        Full model (all tools)       & \textcolor{black}{66.00}             & --                                  & \textcolor{black}{55.20} & --                          \\
        \quad$-$ temporal grounding  & \textcolor{black}{62.00}             & \textcolor{black}{$-4.00$}          & \textcolor{black}{52.80} & \textcolor{black}{$-2.40$}  \\
        \quad$-$ timestamp grounding & \textcolor{black}{61.00}             & \textcolor{black}{$-5.00$}          & \textcolor{black}{52.00} & \textcolor{black}{$-3.20$}  \\
        \quad$-$ spatial grounding   & \textcolor{black}{65.00}             & \textcolor{black}{$-1.00$}          & \textcolor{black}{54.80} & \textcolor{black}{$-0.40$}  \\
        \quad$-$ image search        & \textcolor{black}{58.00}             & \textcolor{black}{$-8.00$}          & \textcolor{black}{48.80} & \textcolor{black}{$-6.40$}  \\
        \quad$-$ web search          & \textcolor{black}{53.00}             & \textcolor{black}{$-13.00$}         & \textcolor{black}{44.00} & \textcolor{black}{$-11.20$} \\
        \quad$-$ page visit          & \textcolor{black}{59.00}             & \textcolor{black}{$-7.00$}          & \textcolor{black}{49.60} & \textcolor{black}{$-5.60$}  \\
        \bottomrule
    \end{tabular}
\end{table}

Removing web search or image search produces the largest declines, consistent with their roles in acquiring external facts and identifying visual entities. Removing spatial grounding has a smaller effect, consistent with its low invocation frequency in Table~\ref{tab:tool_analysis}.

\section{Human--Credit Agreement}
\label{app:human_credit_agreement}

Three annotators independently inspect 50 training tool calls of each type, including the question, relevant video and trajectory context, tool output, target clue, and bypass response. Each annotator assigns every call a credit of 0, 0.5, or 1, and the majority label is used as the human consensus. Table~\ref{tab:human_credit_agreement} reports the agreement between this consensus and the assigned pre-normalization credits.

\begin{table}[H]
    \centering
    \caption{Agreement between human consensus and training-time tool credits by tool type.}
    \label{tab:human_credit_agreement}
    \footnotesize
    \setlength{\tabcolsep}{6pt}
    \resizebox{0.5\textwidth}{!}{%
        \begin{tabular}{lr}
            \toprule
            \textbf{Tool type}  & \textbf{Agreement (\%)} \\
            \midrule
            temporal grounding  & 96.0   \\
            timestamp grounding & 94.0   \\
            spatial grounding   & 92.0   \\
            image search        & 98.0   \\
            web search          & 98.0   \\
            page visit          & 96.0   \\
            \bottomrule
        \end{tabular}}
\end{table}

\clearpage
\section{Prompts}
\label{app:prompts}

\subsection{LLM Judge Prompt}
\label{app:judge_prompt}

We use the following prompt to determine whether a model prediction is semantically consistent with the reference answer. The judge is instructed to focus on factual correctness rather than exact lexical matching and to return a binary decision.

\begin{tcolorbox}[
        enhanced,
        breakable,
        title={Prompt for LLM-Based Answer Evaluation},
        colback=white,
        colframe=black!75,
        colbacktitle=black!75,
        coltitle=white,
        fonttitle=\bfseries\large,
        boxrule=0.8pt,
        arc=2mm,
        left=4mm,
        right=4mm,
        top=3mm,
        bottom=3mm
    ]
    \small
    You are an answer evaluator. Based on the question and the standard answer, determine whether the model answer is correct.

    \medskip
    Allow semantically equivalent wording, abbreviations, formatting differences, and reasonable differences in numerical precision. Mark the answer as incorrect if it contains factual errors, omits essential information, or fails to answer the question.

    \medskip
    \textbf{Question:} \{question\}\\
    \textbf{Standard Answer:} \{standard\_answer\}\\
    \textbf{Model Answer:} \{model\_answer\}

    \medskip
    Output only the following JSON:

    \texttt{\{}\\
    \hspace*{1em}\texttt{"is\_correct": true/false,}\\
    \hspace*{1em}\texttt{"reasoning": "A brief explanation of the judgment"}\\
    \texttt{\}}

    \medskip
    \textbf{Evaluation guidelines:}
    \begin{itemize}[leftmargin=1.5em,itemsep=1pt,topsep=2pt]
        \item Focus on semantic correctness rather than exact wording.
        \item Accept synonyms, abbreviations, alternative names, and reasonable formatting differences.
        \item For numerical answers, allow equivalent units, representations, and reasonable precision differences.
        \item The model answer must contain the essential information required by the question.
        \item Additional information is acceptable unless it contradicts the correct answer.
        \item Mark the answer as incorrect if it contains a factual error, refers to a different entity or concept, omits critical information, or does not answer the question.
        \item If the standard answer itself contains an explanation, compare the core conclusion and required facts rather than requiring identical detail.
    \end{itemize}

    Return only valid JSON in the following format:

    \texttt{\{}\\
    \hspace*{1em}\texttt{"is\_correct": true/false,}\\
    \hspace*{1em}\texttt{"reasoning": "A concise explanation of why the answer is correct or incorrect."}\\
    \texttt{\}}
\end{tcolorbox}

\clearpage
\subsection{Visual Anchor Extraction Prompt}
\label{app:visual_anchor_prompt}

The following prompt is used to extract video-grounded entities and events that can serve as visual anchors for subsequent question construction and external retrieval. The ASR transcript is used only to name or disambiguate targets that are visibly grounded in the video.

\begin{tcolorbox}[
        enhanced,breakable,title={Prompt for Visual Anchor Extraction},
        colback=white,colframe=black!75,colbacktitle=black!75,coltitle=white,
        fonttitle=\bfseries\large,boxrule=0.8pt,arc=2mm,
        left=4mm,right=4mm,top=3mm,bottom=3mm
    ]
    \small
    Watch the video carefully and use the ASR transcript only to help name or disambiguate visible targets.

    \medskip
    \textbf{ASR transcript:} \{asr\_text\}

    \medskip
    Extract video-grounded entities and events. Ignore loose ASR-only mentions and ASR/VSR timestamps.

    \medskip
    \textbf{Rules:}
    \begin{itemize}[leftmargin=1.5em,itemsep=1pt,topsep=2pt]
        \item Every item must include at least one concrete video locator based on visual attributes, readable text, spatial relations, or temporal evidence.
        \item Extract high-value visible targets that can support downstream multi-hop QA with external source facts and can be located again for OCR, image search, Web retrieval, temporal grounding, spatial reasoning, or entity-oriented QA construction.
        \item Scan the entire video for useful anchors, including products, packages, books, posters, signs, screen items, devices, furniture, clothing designs, artworks, places, and named public entities.
        \item Prefer targets with readable text, a logo, model or title information, distinctive design, public-facing identity, clear spatial relations, temporal interactions, or state changes.
        \item For people, retain only named or public-facing individuals, people identified by readable on-screen information, or participants required to localize a useful event.
        \item Set \texttt{identity\_status} to \texttt{known} only when the exact identity is supported by ASR or unambiguous on-screen text. Use \texttt{partially\_resolved} when only a brand, line, coarse name, or title fragment is available, and \texttt{unknown} when only the generic visual category is known.
        \item Do not infer an exact subtype, model, variant, title, or identity solely from visual familiarity. Leave such identification to subsequent OCR or image search.
        \item Drop targets that cannot be localized, cannot be cropped meaningfully, are too weakly visible, or constitute irrelevant background clutter. Avoid near-duplicates and retain at most six unknown or partially resolved entities.
        \item Extract representative visible events, including actions, procedures, state changes, interactions, maneuvers, outcomes, and causal or trigger-like moments. Event descriptions must be based on visible evidence rather than dialogue or external plot knowledge.
        \item If the exact name of an action or event is uncertain, leave the name empty and mark it as requiring expert identification.
    \end{itemize}

    \textbf{Return valid JSON only:}

    \texttt{\{}\\
    \hspace*{1em}\texttt{"visually\_searchable\_entities": [\{}\\
    \hspace*{2em}\texttt{"entity": "...", "category": "...",}\\
    \hspace*{2em}\texttt{"identity\_status": "known | partially\_resolved | unknown",}\\
    \hspace*{2em}\texttt{"exact\_name": "...", "identity\_need": "...",}\\
    \hspace*{2em}\texttt{"visual\_locator": "...", "temporal\_locator": "...",}\\
    \hspace*{2em}\texttt{"spatial\_locator": "...", "asr\_evidence": "..."}\\
    \hspace*{1em}\texttt{\}],}\\
    \hspace*{1em}\texttt{"video\_events": [\{}\\
    \hspace*{2em}\texttt{"event": "...", "event\_status": "recognized | needs\_expert\_name",}\\
    \hspace*{2em}\texttt{"visual\_locator": "...", "observed\_event": "...",}\\
    \hspace*{2em}\texttt{"temporal\_locator": "...", "fine\_grained\_actions": [...],}\\
    \hspace*{2em}\texttt{"asr\_evidence": "..."}\\
    \hspace*{1em}\texttt{\}]}\\
    \texttt{\}}
\end{tcolorbox}

\clearpage
\subsection{QA Generation Prompt}
\label{app:qa_generation_prompt}

Given the extracted visual anchors and source-backed paths, the following prompt first constructs questions and reference answers that require both video evidence and external knowledge. We omit the in-prompt examples for brevity while retaining the task definition, principal constraints, and output schema.

\begin{tcolorbox}[
        enhanced,breakable,title={Prompt for QA Generation},
        colback=white,colframe=black!75,colbacktitle=black!75,coltitle=white,
        fonttitle=\bfseries\large,boxrule=0.8pt,arc=2mm,
        left=4mm,right=4mm,top=3mm,bottom=3mm
    ]
    \small
    Read the source-backed video path payload carefully and create exactly \{num\_qa\} natural, high-quality multi-hop QA records whose answers require combining visual evidence from the video with the provided external source facts.

    \medskip
    \textbf{Input payload:} \{source\_payload\}

    \medskip
    Use \texttt{source\_facts\_for\_answer} as the external information supporting the final answer. The final answer must come from an explicit fact item or be computed directly from explicit facts. Prefer rare facts over common biographical or brand facts.

    \medskip
    \textbf{Question requirements:}
    \begin{itemize}[leftmargin=1.5em,itemsep=1pt,topsep=2pt]
        \item Each question must be a single fluent natural-language question and must require both video evidence and external source facts.
        \item Use indirect visual or temporal descriptions instead of explicitly naming the target entities or events. Do not reveal the final answer or copy the decisive source fact into the question.
        \item Do not mention tools or operations such as timestamps, bounding boxes, grounding, frames, or Web search.
        \item Do not rely on dialogue, speech, captions, subtitles, transcripts, commentary, music, or lyrics.
        \item Event questions must depend on an action, state change, event order, interaction, procedure, outcome, or causal relation unfolding over time.
        \item Entity questions must require identifying one or more visual targets before retrieving external information related to them.
    \end{itemize}

    \textbf{Return valid JSON only:}

    \texttt{\{}\\
    \hspace*{1em}\texttt{"qa\_records": [\{}\\
    \hspace*{2em}\texttt{"qa\_kind": "event | entity",}\\
    \hspace*{2em}\texttt{"source\_path\_ids": ["..."],}\\
    \hspace*{2em}\texttt{"question": "...", "final\_answer": "...",}\\
    \hspace*{2em}\texttt{"answer\_sources": [...]}\\
    \hspace*{1em}\texttt{\}]}\\
    \texttt{\}}
\end{tcolorbox}

\clearpage
\subsection{Initial Trajectory Generation Prompt}
\label{app:initial_trajectory_generation_prompt}

The generated QA records are then paired with initial executable research trajectories. The following prompt determines the grounding, retrieval, and reasoning operations needed to obtain the video and external evidence supporting each answer.

\begin{tcolorbox}[
        enhanced,breakable,title={Prompt for Initial Trajectory Generation},
        colback=white,colframe=black!75,colbacktitle=black!75,coltitle=white,
        fonttitle=\bfseries\large,boxrule=0.8pt,arc=2mm,
        left=4mm,right=4mm,top=3mm,bottom=3mm
    ]
    \small
    Generate an executable reasoning trajectory for each QA record using the source-backed video path payload.

    \medskip
    \textbf{Input QA records:} \{qa\_records\}

    \textbf{Input payload:} \{source\_payload\}

    \medskip
    \textbf{Trajectory requirements:}
    \begin{itemize}[leftmargin=1.5em,itemsep=1pt,topsep=2pt]
        \item Use only \texttt{temporal\_grounding}, \texttt{spatio\_grounding}, \texttt{image\_search}, \texttt{web\_search}, and \texttt{reasoning}.
        \item A visible entity identity must be obtained through \texttt{spatio\_grounding} followed by \texttt{image\_search}. OCR is not represented as a standalone tool action; when a readable clue is required, first localize the relevant region with \texttt{spatio\_grounding} and record the recognized text in the subsequent \texttt{reasoning} step.
        \item Entity records must include the visual grounding chains required to identify their target entities.
        \item Each Web query may use only names or terms already present in the question or produced by earlier steps. It must not contain the answer that the same step is intended to discover.
        \item Use \texttt{reasoning} to combine, compare, select, count, or calculate from previously obtained evidence, and to record OCR evidence from a preceding spatially grounded region. Do not use it to introduce other new video observations or external facts.
        \item Every step must depend on the question or earlier trajectory outputs. Avoid redundant tools and repeated grounding operations.
    \end{itemize}

    \textbf{Return valid JSON only:}

    \texttt{\{}\\
    \hspace*{1em}\texttt{"qa\_records": [\{}\\
    \hspace*{2em}\texttt{"qa\_kind": "event | entity",}\\
    \hspace*{2em}\texttt{"source\_path\_ids": ["..."],}\\
    \hspace*{2em}\texttt{"question": "...", "final\_answer": "...",}\\
    \hspace*{2em}\texttt{"reasoning\_trajectory": [\{}\\
    \hspace*{3em}\texttt{"step": 1, "tool": "temporal\_grounding", "temporal\_target": "..."}\\
    \hspace*{2em}\texttt{\}],}\\
    \hspace*{2em}\texttt{"answer\_sources": [...]}\\
    \hspace*{1em}\texttt{\}]}\\
    \texttt{\}}
\end{tcolorbox}

\clearpage
\subsection{Trajectory Refinement Prompt}
\label{app:trajectory_refinement_prompt}

Before executing an initial trajectory, we refine its tool dependencies and evidence flow using the following prompt. This stage preserves the question and answer while ensuring that every operation is supported by information available at that point in the trajectory.

\begin{tcolorbox}[
        enhanced,title={Prompt for Initial Trajectory Refinement},
        colback=white,colframe=black!75,colbacktitle=black!75,coltitle=white,
        fonttitle=\bfseries\large,boxrule=0.8pt,arc=2mm,
        left=4mm,right=4mm,top=3mm,bottom=3mm
    ]
    \small
    You are a meticulous grounded QA dataset editor. Watch the video frames and refine the existing QA records before downstream temporal and spatial grounding.

    \medskip
    \textbf{Input QA records:} \{qa\_records\}

    \medskip
    \textbf{Rules:}
    \begin{itemize}[leftmargin=1.5em,itemsep=1pt,topsep=2pt]
        \item Preserve \texttt{qa\_kind}, \texttt{evidence\_intent}, question, and final answer unless the record is internally inconsistent. Rewrite only the reasoning trajectory so that it is complete, ordered, and traceable.
        \item Every concrete fact used in a query or reasoning step must come from the question or an earlier trajectory step. Do not introduce hidden names, years, places, events, products, or people in Web queries.
        \item Use only \texttt{temporal\_grounding}, \texttt{spatio\_grounding}, \texttt{image\_search}, \texttt{web\_search}, and \texttt{reasoning}.
        \item If a later step uses a visible entity identity, first apply \texttt{spatio\_grounding} followed by \texttt{image\_search}. OCR is not retained as a standalone tool action; readable text from a spatially grounded region must instead be incorporated into the subsequent \texttt{reasoning} step.
        \item When different visual targets appear at different moments, temporally localize each moment before spatial grounding. The visual query for image search must match the preceding spatial target.
        \item Each Web query may use only names or terms already present in the question or earlier outputs. If additional disambiguating context is required, introduce an earlier retrieval step that explicitly obtains it.
        \item Use reasoning to combine, compare, select, or calculate from evidence produced by earlier steps, and to record OCR evidence from a preceding spatially grounded region. Add Web search before any new external factual claim.
        \item Omit unnecessary tools, keep search queries concise, and renumber all steps from 1 in logical order.
    \end{itemize}

    \textbf{Return valid JSON only:}

    \texttt{\{}\\
    \hspace*{1em}\texttt{"qa\_records": [\{}\\
    \hspace*{2em}\texttt{"qa\_kind": "event | entity",}\\
    \hspace*{2em}\texttt{"question": "...", "evidence\_intent": "...",}\\
    \hspace*{2em}\texttt{"reasoning\_trajectory": [...],}\\
    \hspace*{2em}\texttt{"final\_answer": "...", "why\_long\_horizon": "..."}\\
    \hspace*{1em}\texttt{\}]}\\
    \texttt{\}}
\end{tcolorbox}

\clearpage
\subsection{Model-Conditioned Tool Necessity Filtering Prompts}
\label{app:necessity_prompts}

Tool necessity is evaluated against the capabilities of the base model. For a tool or consecutive tool chain, we withhold its output and ask the model to recover the downstream target directly from the information available before execution. The tool is removed only when the direct prediction is semantically consistent with the expected result.

\begin{tcolorbox}[
        enhanced,title={Prompts for Model-Conditioned Tool Necessity Filtering},
        colback=white,colframe=black!75,colbacktitle=black!75,coltitle=white,
        fonttitle=\bfseries\large,boxrule=0.8pt,arc=2mm,
        left=4mm,right=4mm,top=3mm,bottom=3mm
    ]
    \small
    \textbf{Temporal or timestamp grounding.}\\
    Watch the full video and find one clear timestamp where the following visual target is visible: \{visual\_target\}. Return only JSON containing a timestamp in \texttt{MM:SS.ff} format and a brief description of the visible evidence. If the target is not visible, return an empty timestamp.

    \medskip
    \textbf{Spatial grounding.}\\
    Look at the provided frame and locate the following visual target: \{visual\_target\}. Return only JSON containing a tight bounding box \texttt{[x\_min, y\_min, x\_max, y\_max]} and a brief description of the visible evidence. If the target is not visible, return an empty bounding box.

    \medskip
    \textbf{Image search.}\\
    Look at the provided image crop and answer what visible entity it shows. Return only JSON in the form \texttt{\{"answer": "concise entity name"\}}.

    \medskip
    \textbf{Web search.}\\
    Answer the following query using the model's internal knowledge without invoking Web search: \{query\}. Return only JSON in the form \texttt{\{"answer": "concise answer"\}}.

    \medskip
    \textbf{Consistency decision.}\\
    Given \texttt{model\_answer} and \texttt{expected\_answer}, determine whether they are semantically consistent. Return only JSON:

    \texttt{\{"consistent": true/false, "reason": "short reason"\}}.
\end{tcolorbox}

\clearpage
\subsection{Reflection and Retry Prompts}
\label{app:retry_prompts}

When execution produces insufficient or unreliable evidence, the pipeline invokes a targeted retry according to the failed operation. The following prompts show the two principal recovery behaviors: revising video localization and rewriting an external search query.

\begin{tcolorbox}[
        enhanced,title={Prompt for Grounding Retry},
        colback=white,colframe=black!75,colbacktitle=black!75,coltitle=white,
        fonttitle=\bfseries\large,boxrule=0.8pt,arc=2mm,
        left=4mm,right=4mm,top=3mm,bottom=3mm
    ]
    \small
    The previous grounding attempt did not provide a crop that clearly contains the required visual target.

    \medskip
    \textbf{Visual target:} \{visual\_target\}\\
    \textbf{Previous attempt:} \{previous\_attempt\}\\
    \textbf{Failure reason:} \{issue\}

    \medskip
    Re-examine the available video evidence and determine how the target should be localized again. If the target appears at a different moment, propose a revised temporal description for locating a clearer timestamp. If the timestamp is appropriate but the crop is inaccurate, propose a tighter and more precise spatial description. Do not introduce an entity identity that has not been established by visible evidence.

    \medskip
    Return only valid JSON describing the retry action, the revised grounding target, and a brief reason.
\end{tcolorbox}

\begin{tcolorbox}[
        enhanced,title={Prompt for Web Query Rewrite},
        colback=white,colframe=black!75,colbacktitle=black!75,coltitle=white,
        fonttitle=\bfseries\large,boxrule=0.8pt,arc=2mm,
        left=4mm,right=4mm,top=3mm,bottom=3mm
    ]
    \small
    Rewrite the Web search query so the next search is more likely to return the expected information.

    \medskip
    \textbf{Original query:} \{original\_query\}\\
    \textbf{Expected information:} \{expected\_answer\}\\
    \textbf{Available context:} \{available\_context\}\\
    \textbf{Previous attempts:} \{previous\_attempts\}

    \medskip
    \textbf{Rules:}
    \begin{itemize}[leftmargin=1.5em,itemsep=1pt,topsep=2pt]
        \item Keep the rewritten query concise.
        \item Include the key entity or topic and the missing attribute to verify.
        \item Do not include the expected answer itself in the rewritten query.
        \item Use only information from the available context, original query, or previous attempts. Do not add facts that this search step is intended to discover.
    \end{itemize}

    Return only valid JSON:

    \texttt{\{"query": "concise rewritten query", "reason": "brief reason for the rewrite"\}}.
\end{tcolorbox}